\PassOptionsToPackage{table}{xcolor}
\documentclass[10pt,twocolumn,letterpaper]{article}

\usepackage{multirow}
\usepackage{algorithm}
\usepackage{algpseudocode}
\usepackage{bbm}
\usepackage{float}
\usepackage{graphicx}
\usepackage{circledsteps}
\usepackage[pagenumbers]{cvpr} 

\definecolor{blue}{rgb}{0.21,0.49,0.74}
\usepackage[pagebackref,breaklinks,colorlinks,citecolor=blue]{hyperref}

\def\Approach{Open3DIS}

\definecolor{mydarkblue}{rgb}{0,0.08,1}
\definecolor{mydarkgreen}{rgb}{0.02,0.6,0.02}
\definecolor{myred}{rgb}{1.0,0.0,0.0}
\definecolor{myred2}{rgb}{0.7,0.1,0.1}
\definecolor{mydarkblue2}{rgb}{0.05,0.1,0.7}
\definecolor{mypurple}{rgb}{111,0,255}
\definecolor{mypurple2}{rgb}{111,0,111}

\newcommand\blfootnote[1]{%
  \begingroup
\renewcommand\thefootnote{}\footnote{#1}%
  \addtocounter{footnote}{-1}%
  \endgroup
}

\title{\Approach: Open-Vocabulary 3D Instance Segmentation with 2D Mask Guidance}

\author{
Phuc Nguyen$^{1*}$ \qquad Tuan Duc Ngo$^{1,4*}$ \qquad Evangelos Kalogerakis$^{4}$  \\
Chuang Gan$^{2,4}$ \quad Anh Tran$^{1}$ \qquad Cuong Pham$^{1,3}$ \qquad Khoi Nguyen$^{1}$ \\
\normalsize{$^1$VinAI Research \qquad $^2$MIT-IBM Watson AI Lab \qquad $^3$Posts \& Telecommunications Inst. of Tech. \qquad $^4$UMass Amherst } \\
\tt\small \{v.phucnda,  v.anhtt152, v.khoindm\}@vinai.io \qquad \{tdngo, kalo\}@cs.umass.edu \\
\tt\small \qquad ganchuang@csail.mit.edu \tt\small \qquad cuongpv@ptit.edu.vn  \\
\url{https://open3dis.github.io/}
}

\begin{document}

\def\mA{\mathcal{A}}
\def\mB{\mathcal{B}}
\def\mC{\mathcal{C}}
\def\mD{\mathcal{D}}
\def\mE{\mathcal{E}}
\def\mF{\mathcal{F}}
\def\mG{\mathcal{G}}
\def\mH{\mathcal{H}}
\def\mI{\mathcal{I}}
\def\mJ{\mathcal{J}}
\def\mK{\mathcal{K}}
\def\mL{\mathcal{L}}
\def\mM{\mathcal{M}}
\def\mN{\mathcal{N}}
\def\mO{\mathcal{O}}
\def\mP{\mathcal{P}}
\def\mQ{\mathcal{Q}}
\def\mR{\mathcal{R}}
\def\mS{\mathcal{S}}
\def\mT{\mathcal{T}}
\def\mU{\mathcal{U}}
\def\mV{\mathcal{V}}
\def\mW{\mathcal{W}}
\def\mX{\mathcal{X}}
\def\mY{\mathcal{Y}}
\def\mZ{\mathcal{Z}} 

\def\bbN{\mathbb{N}} 
\def\bbR{\mathbb{R}} 
\def\bbP{\mathbb{P}} 
\def\bbQ{\mathbb{Q}} 
\def\bbE{\mathbb{E}}

\def\1n{\mathbf{1}_n}
\def\0{\mathbf{0}}
\def\1{\mathbf{1}}

\def\A{{\bf A}}
\def\B{{\bf B}}
\def\C{{\bf C}}
\def\D{{\bf D}}
\def\E{{\bf E}}
\def\F{{\bf F}}
\def\G{{\bf G}}
\def\H{{\bf H}}
\def\I{{\bf I}}
\def\J{{\bf J}}
\def\K{{\bf K}}
\def\L{{\bf L}}
\def\M{{\bf M}}
\def\N{{\bf N}}
\def\O{{\bf O}}
\def\P{{\bf P}}
\def\Q{{\bf Q}}
\def\R{{\bf R}}
\def\S{{\bf S}}
\def\T{{\bf T}}
\def\U{{\bf U}}
\def\V{{\bf V}}
\def\W{{\bf W}}
\def\X{{\bf X}}
\def\Y{{\bf Y}}
\def\Z{{\bf Z}}

\def\a{{\bf a}}
\def\b{{\bf b}}
\def\c{{\bf c}}
\def\d{{\bf d}}
\def\e{{\bf e}}
\def\f{{\bf f}}
\def\g{{\bf g}}
\def\h{{\bf h}}
\def\i{{\bf i}}
\def\j{{\bf j}}
\def\k{{\bf k}}
\def\l{{\bf l}}
\def\m{{\bf m}}
\def\n{{\bf n}}
\def\o{{\bf o}}
\def\p{{\bf p}}
\def\q{{\bf q}}
\def\r{{\bf r}}
\def\s{{\bf s}}
\def\t{{\bf t}}
\def\u{{\bf u}}
\def\v{{\bf v}}
\def\w{{\bf w}}
\def\x{{\bf x}}
\def\y{{\bf y}}
\def\z{{\bf z}}

\def\balpha{\mbox{\boldmath{$\alpha$}}}
\def\bbeta{\mbox{\boldmath{$\beta$}}}
\def\bdelta{\mbox{\boldmath{$\delta$}}}
\def\bgamma{\mbox{\boldmath{$\gamma$}}}
\def\blambda{\mbox{\boldmath{$\lambda$}}}
\def\bsigma{\mbox{\boldmath{$\sigma$}}}
\def\btheta{\mbox{\boldmath{$\theta$}}}
\def\bomega{\mbox{\boldmath{$\omega$}}}
\def\bxi{\mbox{\boldmath{$\xi$}}}
\def\bnu{\mbox{\boldmath{$\nu$}}}                                  
\def\bphi{\mbox{\boldmath{$\phi$}}}
\def\bmu{\mbox{\boldmath{$\mu$}}}

\def\bDelta{\mbox{\boldmath{$\Delta$}}}
\def\bOmega{\mbox{\boldmath{$\Omega$}}}
\def\bPhi{\mbox{\boldmath{$\Phi$}}}
\def\bLambda{\mbox{\boldmath{$\Lambda$}}}
\def\bSigma{\mbox{\boldmath{$\Sigma$}}}
\def\bGamma{\mbox{\boldmath{$\Gamma$}}}
                                  
\newcommand{\myprob}[1]{\mathop{\mathbb{P}}_{#1}}

\newcommand{\myexp}[1]{\mathop{\mathbb{E}}_{#1}}

\newcommand{\mydelta}[1]{1_{#1}}

\newcommand{\myminimum}[1]{\mathop{\textrm{minimum}}_{#1}}
\newcommand{\mymaximum}[1]{\mathop{\textrm{maximum}}_{#1}}    
\newcommand{\mymin}[1]{\mathop{\textrm{minimize}}_{#1}}
\newcommand{\mymax}[1]{\mathop{\textrm{maximize}}_{#1}}
\newcommand{\mymins}[1]{\mathop{\textrm{min.}}_{#1}}
\newcommand{\mymaxs}[1]{\mathop{\textrm{max.}}_{#1}}  
\newcommand{\myargmin}[1]{\mathop{\textrm{argmin}}_{#1}} 
\newcommand{\myargmax}[1]{\mathop{\textrm{argmax}}_{#1}} 
\newcommand{\myst}{\textrm{s.t. }}

\newcommand{\denselist}{\itemsep -1pt}
\newcommand{\sparselist}{\itemsep 1pt}

\definecolor{pink}{rgb}{0.9,0.5,0.5}
\definecolor{purple}{rgb}{0.5, 0.4, 0.8}   
\definecolor{gray}{rgb}{0.3, 0.3, 0.3}
\definecolor{mygreen}{rgb}{0.2, 0.6, 0.2}

\newcommand{\cyan}[1]{\textcolor{cyan}{#1}}
\newcommand{\red}[1]{\textcolor{red}{#1}}  
\newcommand{\blue}[1]{\textcolor{blue}{#1}}
\newcommand{\magenta}[1]{\textcolor{magenta}{#1}}
\newcommand{\pink}[1]{\textcolor{pink}{#1}}
\newcommand{\green}[1]{\textcolor{green}{#1}} 
\newcommand{\gray}[1]{\textcolor{gray}{#1}}    
\newcommand{\mygreen}[1]{\textcolor{mygreen}{#1}}    
\newcommand{\purple}[1]{\textcolor{purple}{#1}}       

\definecolor{greena}{rgb}{0.4, 0.5, 0.1}
\newcommand{\greena}[1]{\textcolor{greena}{#1}}

\definecolor{bluea}{rgb}{0, 0.4, 0.6}
\newcommand{\bluea}[1]{\textcolor{bluea}{#1}}
\definecolor{reda}{rgb}{0.6, 0.2, 0.1}
\newcommand{\reda}[1]{\textcolor{reda}{#1}}

\def\changemargin#1#2{\list{}{\rightmargin#2\leftmargin#1}\item[]}
\let\endchangemargin=\endlist
                                               
\newcommand{\cm}[1]{}

\newcommand{\mhoai}[1]{{\color{magenta}\textbf{[MH: #1]}}}

\newcommand{\mtodo}[1]{{\color{red}$\blacksquare$\textbf{[TODO: #1]}}}
\newcommand{\myheading}[1]{\vspace{1ex}\noindent \textbf{#1}}
\newcommand{\htimesw}[2]{\mbox{$#1$$\times$$#2$}}


\newif\ifshowsolution
\showsolutiontrue

\ifshowsolution  
\newcommand{\Solution}[2]{\paragraph{\bf $\bigstar $ SOLUTION:} {\sf #2} }
\newcommand{\Mistake}[2]{\paragraph{\bf $\blacksquare$ COMMON MISTAKE #1:} {\sf #2} \bigskip}
\else
\newcommand{\Solution}[2]{\vspace{#1}}
\fi

\newcommand{\truefalse}{
\begin{enumerate}
	\item True
	\item False
\end{enumerate}
}

\newcommand{\yesno}{
\begin{enumerate}
	\item Yes
	\item No
\end{enumerate}
}

\newcommand{\Sref}[1]{Sec.~\ref{#1}}
\newcommand{\Eref}[1]{Eq.~(\ref{#1})}
\newcommand{\Fref}[1]{Fig.~\ref{#1}}
\newcommand{\Tref}[1]{Table~\ref{#1}}


\twocolumn[{
\renewcommand\twocolumn[1][]{#1}%
\maketitle
\vspace{-20pt}
\begin{center}%
\includegraphics[width=.85\linewidth]{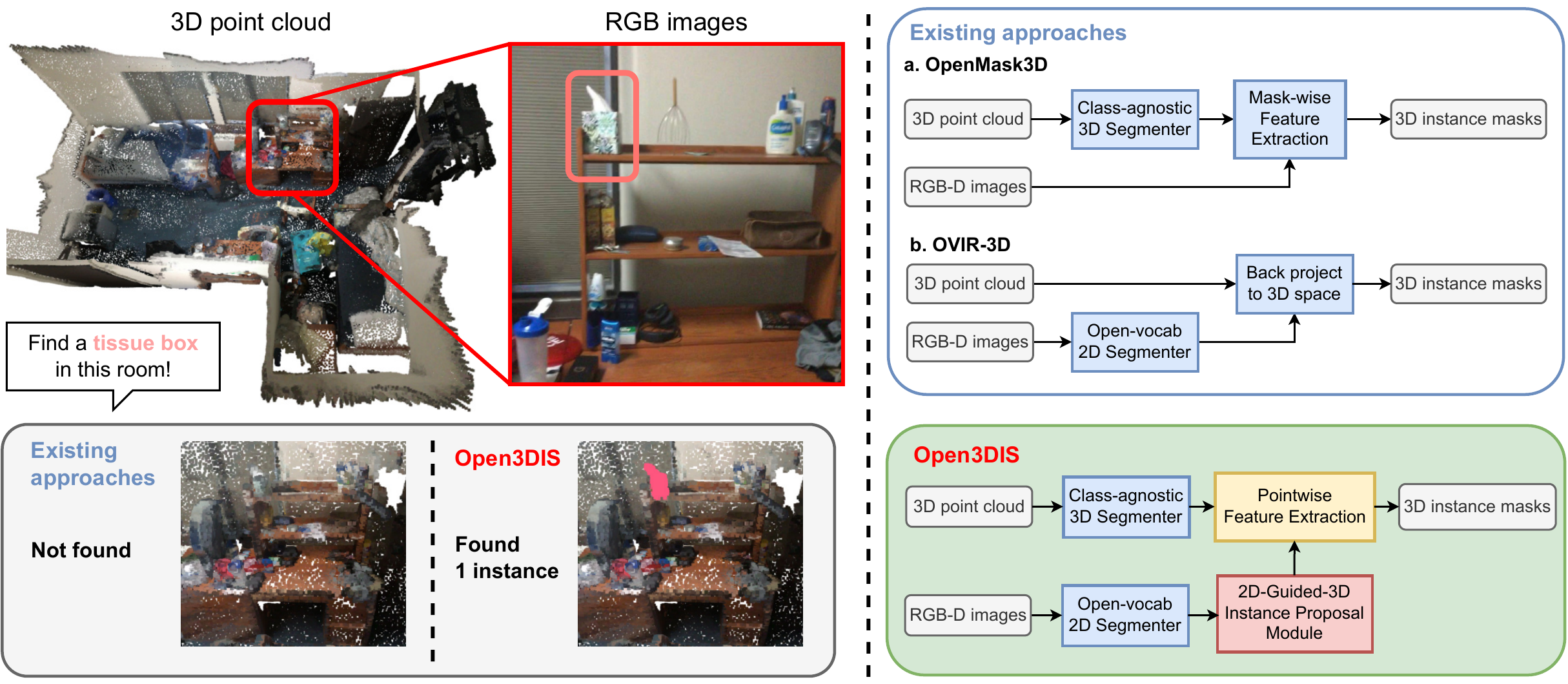}
\vspace{-5pt}
\captionof{figure}{\textbf{Left:} While leading open-vocabulary 3D instance segmentation methods like OpenMask3D \cite{openmask3d} and OVIR-3D \cite{ovir3d} often struggle with small or ambiguous instances, particularly those from uncommon classes, \Approach~excels in segmenting such cases. It outperforms existing methods by about ${\sim}1.5\x$ in average precision on ScanNet200 \cite{scannet200}. \textbf{Right:} 
\Approach~aggregates proposals from both point cloud-based instance segmenters and 2D image-based networks. Our method incorporates novel components (red and yellow boxes) that perform
aggregation and mapping of 2D masks to the point cloud across multiple frames, as well as 3D-aware feature extraction for effectively comparing object proposals to text queries.}
\label{fig:teaser}
\end{center}
}]

\begin{abstract}
\vspace{-6pt}
We introduce \Approach, a novel solution designed to tackle the problem of Open-Vocabulary Instance Segmentation within 3D scenes. Objects within 3D environments exhibit diverse shapes, scales, and colors, making precise instance-level identification a challenging task. Recent advancements in Open-Vocabulary scene understanding have made significant strides in this area by employing class-agnostic 3D instance proposal networks for object localization and learning queryable features for each 3D mask. While these methods produce high-quality instance proposals, they struggle with identifying small-scale and geometrically ambiguous objects. The key idea of our method is a new module that aggregates 2D instance masks across frames and maps them to geometrically coherent point cloud regions as high-quality object proposals addressing the above limitations. These are then combined with 3D class-agnostic instance proposals to include a wide range of objects in the real world. 
To validate our approach, we conducted experiments on three prominent datasets, including ScanNet200, S3DIS, and Replica, demonstrating significant performance gains in segmenting objects with diverse categories over the state-of-the-art approaches. 
\end{abstract}
\vspace{-14pt} 

\section{Introduction}
\noindent This paper\blfootnote{$^*$: Equal contribution}  addresses the challenging problem of open-vocabulary 3D point cloud instance segmentation (OV-3DIS). Given a 3D scene represented by a point cloud, we seek to obtain a set of binary instance masks of any classes of interest, which may not exist during the training phase. This problem arises to overcome the inherent constraints of the conventional fully supervised 3D instance segmentation (3DIS) approaches \cite{yi2019gspn, zhang2021point,he2021dyco3d,he2022pointinst3d,sun2022superpoint, vu2022softgroup,ngo2023isbnet, Schult23ICRA}, which are bound by a closed-set framework -- restricting recognition to a predefined set of object classes that are determined by the training datasets. This task has a wide range of applications in robotics and VR systems. This capability can empower robots or agents to identify and localize objects of any kind in a 3D environment using textual descriptions that detail names, appearances, functionalities, and more.





There are a few studies addressing the OV-3DIS so far \cite{pla,ding2023lowis3d,openmask3d,ovir3d}. Most recently, \cite{openmask3d} proposes the use of a pre-trained 3DIS model instance proposals network to capture the geometrical structure of 3D point cloud scenes and generate high-quality instance masks. However, this approach faces challenges in recognizing rare objects due to their incomplete appearance in the 3D point cloud scene and the limited detection capabilities of pre-trained 3D models for such infrequent classes. Another approach involves leveraging 2D off-the-shelf open-vocabulary understanding models \cite{ovir3d, sam3d} to easily capture novel classes. Nevertheless, translating these 2D proposals from images to 3D point cloud scenes is a challenging task. This is because of the fact that 2D proposals capture only the visible portions of 3D objects and may also include irrelevant 
regions, such as the background. These two approaches are summarized in Fig.~\ref{fig:teaser}.

In this work, we introduce \Approach, a method for OV-3DIS that extends the understanding capability beyond predefined concept sets. Given an RGB-D sequence of images and the corresponding 3D reconstructed point cloud scene, \Approach~addresses the limitations of existing approaches. It complements two sources of 3D instance proposals by employing a 3D instance network and a 2D-guide-3D Instance Proposal Module to achieve sufficient 3D object binary instance masks. The module (our key contribution) extracts geometrically coherent regions from the point cloud under the guidance of 2D predicted masks across multiple frames and aggregates them into higher-quality 3D proposals. Later, Pointwise Feature Extraction aggregates CLIP features for each instance in a multi-scale manner across multiple views, constructing instance-aware point cloud features for open-vocabulary instance segmentation. 

To assess the open-vocabulary capability of \Approach, we conduct experiments on the ScanNet200 \cite{scannet200}, S3DIS \cite{armeni20163d}, and Replica \cite{straub2019replica} datasets. \Approach~achieves state-of-the-art results in OV-3DIS, surpassing prior works by a significant margin. 
Especially, \Approach~delivers a noteworthy performance improvement of ${\sim}1.5$ times compared to the leading method on the large-scale dataset ScanNet200.

In summary, the contributions of our work are as follows:
\begin{enumerate}[leftmargin=7.5mm]
    
    \item We present the ``2D-Guided 3D Proposal Module'' creating precise 3D proposals by clustering cohesive point cloud regions using aggregated 2D instance masks from multi-view RGB-D images.
    \item We introduce a novel pointwise feature extraction method for open-vocabulary 3D object proposals.
    \item \Approach~achieves state-of-the-art results on ScanNet200, S3DIS, and Replica datasets, exhibiting comparable performance to fully supervised methods.
\end{enumerate}

\section{Related Work}
\label{sec:relatedwork}

\begin{figure*}[t]
  \centering
  \includegraphics[width=0.85\linewidth]{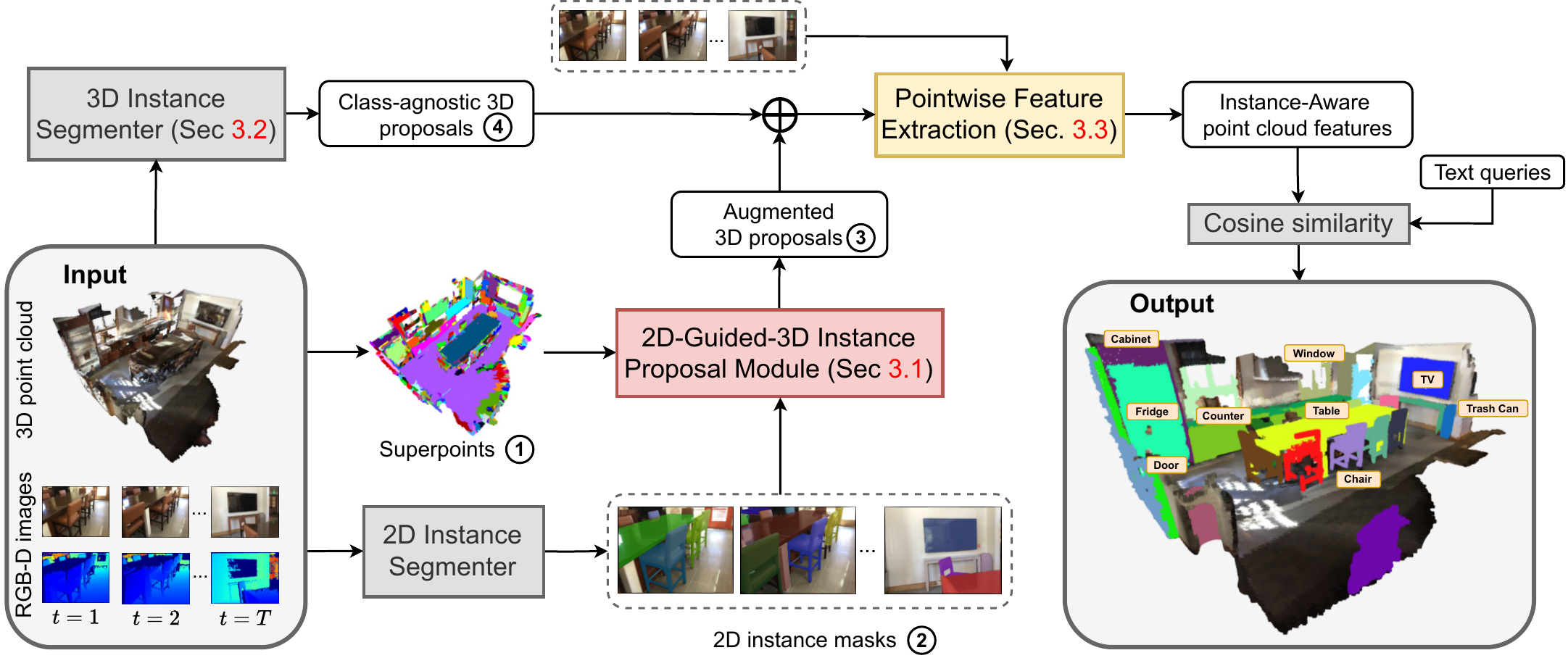}
   \vspace{-7pt}
   \caption{\textbf{Overview of \Approach}. A pre-trained class-agnostic 3D Instance Segmenter proposes initial 3D objects, while a 2D Instance Segmenter generates masks for video frames. Our 2D-Guided-3D Instance Proposal Module (Sec.~\ref{subsec:2D-G-3D-IPM}) combines superpoints and 2D instance masks to enhance 3D proposals, integrating them with the initial 3D proposals. Finally, the Pointwise Feature Extraction module (Sec.~\ref{subsec:PFE}) correlates instance-aware point cloud CLIP features with text embeddings to generate the ultimate instance masks.}
   \vspace{-12pt}
   \label{fig:architecture}
\end{figure*}

\myheading{Open-Vocabulary 2D scene understanding} methods aim to recognize both base and novel classes in testing where the base classes are seen during training while the novel classes are not. Based on the types of recognition tasks, we can categorize them into open-vocabulary object detection (OVOD) \cite{zhong2022regionclip, pham2023lp, liu2023grounding, zang2022open, wang2023object, Kaul23, yao2023detclipv2}, open-vocabulary semantic segmentation (OVSS) \cite{li2022languagedriven, xu2022groupvit, li2023grounded, zou2023segment, wu2023diffumask, ding2021decoupling}, and open-vocabulary instance segmentation (OVIS) \cite{huynh2022open, ding2022open, he2023semantic, vs2023mask, wu2023betrayed, zhang2023simple}. A typical approach for handling the novel classes is to leverage a pre-trained visual-text embedding model, such as CLIP \cite{radford2021clip} or ALIGN \cite{jia2021scaling} as a joint text-image embedding where base and novel classes co-exist, in order to transfer the models' capabilities on base classes to novel classes. However, these methods cannot trivially extend to 3D point clouds because 3D point clouds are unordered and imbalanced in density, and the variance in appearance and shape is much larger than that of 2D images.

\myheading{Fully-Supervised 3D Instance Segmentation (F-3DIS)} aims to segment 3D point cloud into instances of training classes. Methods of F-3DIS can be categorized into three main groups: box-based \cite{hou20193d,yang2019learning,yi2019gspn}, cluster-based \cite{wang2018sgpn,jiang2020pointgroup,chen2021hierarchical,vu2022softgroup,dong2022learning}, and dynamic convolution-based \cite{he2021dyco3d,sun2022superpoint,he2022pointinst3d,wu20223d,Schult23ICRA,liu20223d, ngo2023isbnet} techniques. Box-based methods detect and segment the foreground region inside each 3D proposal box to get instance masks. Cluster-based methods employ the predicted object centroid to group points to clusters or construct a tree or graph structure and subsequently dissect these into subtrees or subgraphs \cite{liang2021instance,hui2022learning}. For the third group, Mask3D \cite{Schult23ICRA} and ISBNet \cite{ngo2023isbnet}, proposed using dynamic convolution whose kernels, representative of different object instances, are convoluted with pointwise features to derive instance masks. In this paper, we use ISBNet as a 3D network, yet with necessary adaptations to output 3D class-agnostic proposals.

\myheading{Open-Vocabulary 3D semantic segmentation (OV-3DSS) and object detection (OV-3DOD)} enable the semantic understanding of 3D scenes in an open-vocabulary manner, including affordances, materials, activities, and properties within unseen environments. 
This capability is highlighted in recent work \cite{openscene, gu2023conceptgraphs, hong20233d} for OV-3DSS and \cite{lu2023open, cao2023coda, zhu2023object2scene} for OV-3DOD. Nevertheless, these methods cannot precisely locate and distinguish 3D objects with 3D instance masks, and thus cannot fully describe 3D object shapes. 

\myheading{Open-Vocabulary 3D instance segmentation (OV-3DIS)} concerns segmenting both seen and unseen classes (during training) of a 3D point cloud into instances. Methods of OV-3DIS can be split into 3 groups: open-vocabulary semantic segmentation-based, text description and 3D proposal contrastive learning based, and 2D open-vocabulary powered approaches. 
\textbf{The first group} includes OpenScene \cite{openscene} and Clip3D \cite{hegde2023clip} utilize clustering techniques such as DBScan on OV-3DSS results to generate 3D instance proposals. However, their quality relies on clustering accuracy and can lead to unreliable results for unseen classes.
On the other hand, \textbf{the second group} comprising PLA \cite{pla}, RegionPLC \cite{yang2023regionplc}, and Lowis3D \cite{ding2023lowis3d} focuses on training the 3D instance proposal network along with a contrastive open-vocabulary between the predicted proposals and their corresponding text captions. However, when growing the number of classes, these methods struggle to handle and may degrade their ability to distinguish diverse object classes.
For \textbf{the final group}, OpenMask3D \cite{openmask3d} utilizes a pre-trained 3DIS model to generate class-agnostic 3D proposals, which are subsequently classified based on their CLIP score from 2D mask projections. Similarly, OpenIns3D \cite{huang2023openins3d} employs a pre-trained 3DIS model and addresses the issue through its Mask-Snap-Lookup module, utilizing synthetic-scene images across multiple scales. However, challenges arise for the pre-trained 3DIS model when identifying small or uncommon object categories with unique geometric structures.
Conversely, OVIR-3D \cite{ovir3d}, SAM3D \cite{sam3d}, SAMPro3D \cite{xu2023sampro3d}, MaskClustering \cite{yan2024maskclustering} and SAI3D \cite{yin2023sai3d} leverage pretrained 2D open-vocabulary models to generate 2D instance masks, which are then back-projected onto the associated 3D point cloud. However, imperfect alignment of the 2D segmentation masks with objects leads to the inclusion of background points in foreground objects, resulting in suboptimal quality of 3D proposals.
Nonetheless, the advantage of this group over other groups is in their leverage of 2D pretrained model on large-scale datasets such as CLIP \cite{radford2021clip} or SAM \cite{kirillov2023segmentanythign} which can be scaled to hundreds of classes as in ScanNet200 \cite{scannet200}.
Following the final group, \Approach~generates high-quality 3D instance proposals by combining 3D masks from a 3DIS network with proposals produced by grouping geometrically coherent regions (superpoints) with the guidance of 2D instance masks.
This complements the class-agnostic 3D instance proposals from 3D networks. Our method excels at capturing rare objects while preserving their 3D geometrical structures, achieving state-of-the-art performance in the OV-3DIS domain.


\section{Method}
\label{sec:method}



Our approach processes a 3D point cloud and an RGB-D sequence, producing a set of 3D binary masks indicating object instances in the scene. We assume known camera parameters for each frame. Our architecture is depicted in Fig.~\ref{fig:architecture}. Similarly to prior work \cite{openmask3d, pla, yang2023regionplc}, we employ a \emph{3DIS network module} to extract object proposals directly from the 3D point cloud. This module leverages 3D convolution and attention mechanisms, capturing spatial and structural relations for robust 3D object instance detection. Despite its advantages, sparse point clouds, sampling artifacts, and noise can lead to missed objects, especially for small objects e.g., the tissue box in Fig.~\ref{fig:teaser}.

Our approach integrates a novel \emph{2D-Guided-3D instance proposal module} , leveraging 2D instance segmentation networks trained on large image datasets to better capture smaller objects in individual images. However, resulting 2D masks may only capture parts of actual 3D object instances due to occlusions (Fig.~\ref{fig:architecture} -  \textbf{\Circled[]{2}}). To address this, we propose a strategy that constructs 3D object instance proposals by hierarchically aggregating and merging point cloud regions from back-projected 2D masks of the same object. To enhance the robustness and geometric homogeneity, we use ``superpoints'' \cite{felzenszwalb2004efficient} during the merging process. This yields complete object instances, complementing those extracted by 3DIS networks.
Detailed analysis in Tab.~\ref{tab:recall_rate} on Scannet200 dataset \cite{scannet200} exhibits the significant enhancement in recall rate, especially for \textit{rare} classes, when integrating 2D and 3D proposals.


To enable open-vocabulary classification, we additionally employ a \emph{point-wise feature extraction module} to construct a dense feature map across the 3D point cloud. In the following sections, we explain our modules in more detail, starting with the 2D-Guided-3D Instance Proposal Module which constitutes our main contribution.  

\begin{table}[t]
\small
\centering
\begin{tabular}{lcccc}
\toprule
 & \textbf{Recall} & \textbf{Recall}$_{\text{head}}$  & \textbf{Recall}$_{\text{com}}$ & \textbf{Recall}$_{\text{tail}}$ \\
\midrule
Only 3D & 61.63 & 81.92 & 53.68 & 12.06  \\
Only 2D & 68.61 & 76.66 & 74.73 & 34.68 \\
\midrule
2D and 3D & 73.29 & 87.48 & 74.16 & 34.31 \\
\bottomrule
\end{tabular}
\vspace{-5pt}
\caption{Recall rate (\%) of 2D, 3D, or combined proposals.}
\label{tab:recall_rate}
\vspace{-15pt}
\end{table}

\subsection{2D-Guided-3D Instance Proposal Module}
\label{subsec:2D-G-3D-IPM}
This module takes as input a 3D point cloud $\P=\{\p_n\}_{n=1}^N$, where $N$ is the number of points, and $\p_i \in \bbR^6$ includes 3D coordinates and RGB color. Additionally, it receives an RGB-D video sequence $\V = \left \{(\I_t,\D_t, \Pi_t) \right \}_{t=1}^T$, where each frame $t$ contains RGB image $\I_t$, depth map $\D_t$, and camera matrix $\Pi_t$ (i.e., the product of intrinsic and extrinsic matrices used for projecting 3D points onto the image plane). The output comprises $K_1$ binary instance masks represented in a $K_1 {\times} N$ binary matrix $\M_1$ (Fig.~\ref{fig:architecture} - \textbf{\Circled[]{3}}).

\myheading{Superpoints.} In a pre-processing step, we utilize the method of \cite{felzenszwalb2004efficient} to group points into geometrically homogeneous regions, termed superpoints (Fig.~\ref{fig:architecture} - \textbf{\Circled[]{1}}). This yields a set of $U$ superpoints $\{\q_u\}_{u=1}^U \in \{0, 1\}^{U \times N}$, where $\q_u$ is a binary mask of points. Superpoints enhance processing efficiency in the later stages of our pipeline and contribute to well-formed candidate object instances.


\myheading{Per-frame superpoint merging.} For all input frames, we utilize a pretrained 2D instance segmenter, employing  Grounding-DINO \cite{liu2023grounding} and SAM \cite{kirillov2023segmentanything}. The network outputs a set of 2D masks (Fig.~\ref{fig:architecture} -  \textbf{\Circled[]{2}}). For each 2D mask with index $m$ (unique across all frames), we calculate the IoU $o_{u,m}$ with each superpoint $\q_u$ when projecting all points of $\q_u$ onto the image plane of mask $m$ using the known camera matrix, excluding points outside the camera's field of view, and determining image pixels containing projected points. A superpoint is considered to have sufficient overlap with a 2D mask if the IoU is higher than a threshold $o_{u,m} > \tau_{iou}$. 


However, 2D masks may include background regions or parts of nearby objects, making IoU alone insufficient to determine superpoints belonging to a 3D proposal. To address this, we leverage the 3D backbone of a 3D proposal network \cite{ngo2023isbnet, Schult23ICRA} to extract per-point feature $\F^{\text{3D}} \in \mathbb{R}^{N \times D^{\text{3D}}}$ and measure feature similarity among these superpoints $\q_u$ whose features are determined by averaging their point features $\f^{\text{3D}}_u \in \mathbb{R}^{1 \times D^{\text{3D}}}$. For each 2D instance mask $\m_i^{\text{2D}}$, we initiate a point cloud region $\r_i$ with the superpoint having the largest IoU with the mask. We extend this region by merging with neighboring superpoints $\q_u$ that meet the overlapping condition ($\tau_{iou}$) and also have the highest cosine similarity $s^{\text{max}}_i = \max_{u' \in \r_i} \text{cos}(\f^{\text{3D}}_{u'}, \f^{\text{3D}}_u)$ with those already in the region $\r_i$ above a threshold ($s_i^{\text{max}} > \tau_{sim}$) (we will discuss the effect of all thresholds in our results section). The growth continues until no other overlapping or neighboring superpoints are found. 
Our superpoint merging procedure, compared to using points alone or other merging strategies (see Tab.~\ref{tab:ablation_2d_mask_guided}), produces more well-formed point cloud regions corresponding to 2D masks per frame.

\begin{figure}[t]
  \centering
  \includegraphics[width=0.98\linewidth]{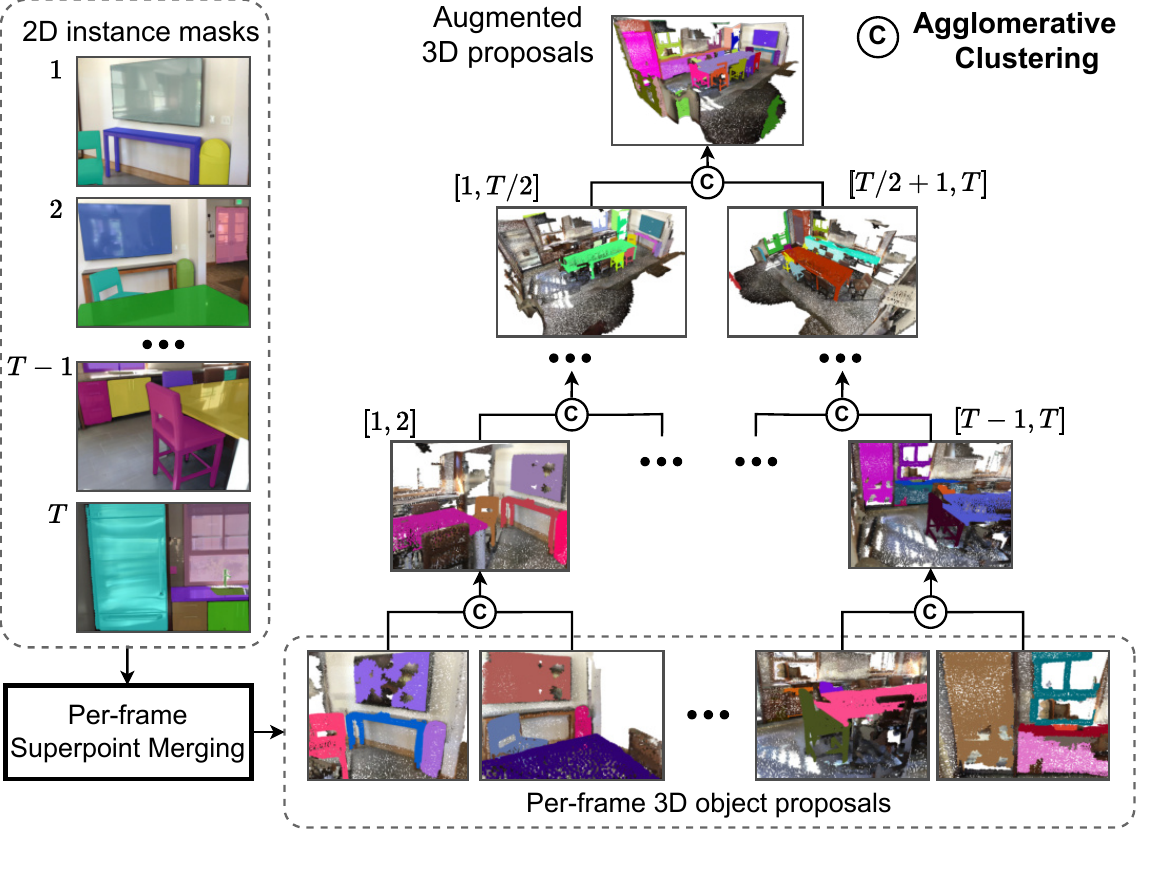}
   \vspace{-10pt}
   \caption{\textbf{2D-Guided-3D Instance Proposal Module.} We generate initial 3D proposals using Per-frame Superpoint Merging, followed by hierarchical traversal across the RGB-D sequence to merge region sets between frames using Agglomerative clustering.
   }\label{fig:2dmaskguide}
   \vspace{-14pt}
\end{figure}

\myheading{3D object proposal formation.} To create 3D object proposals, one option is to utilize the point cloud regions obtained from the merging procedure across individual frames. However, this results in fragmented proposals, capturing only parts of object instances, as the regions correspond to 2D masks from single views (Fig.~\ref{fig:architecture} - \textbf{\Circled[]{2}}). To address this, we merge point cloud regions from different frames in a bottom-up manner, creating more complete and coherent 3D object masks. Agglomerative clustering combines region sets from pairs of frames until no compatible pairs remain.
The resulting set includes merged and standalone regions, which can be matched with other region sets from subsequent frames. In the following paragraphs, we discuss three crucial design choices in this process: (a) the matching score between region pairs, (b) the matching process between sets of regions, and (c) the order of frames or region sets used in matching and merging. 

\myheading{Matching score.} For a pair of point cloud regions
$(\r_i, \r_j)$, 
we define a matching score based on (a) feature similarity and (b) overlap degree. Their feature-based similarity $s'$
is measured through cosine similarity between the regions' feature vectors $\f_i^{\text{3D}}$, or $s'_{i,j}=\text{cos}(\f_i^{\text{3D}}, \f_j^{\text{3D}})$, which are in turn computed as the average of their point features.    
While this measures if the regions belong to the same object's shape, it may yield high similarity for duplicate instances with the same geometry.
To address this, we also consider the degree of overlap, expressed as the IoU $o'_{i,j}=\text{IoU}(\r_i, \r_j)$ between the two regions $\r_i, \r_j$, which is expected to be high for overlapping regions of the same instance. Two regions are considered matching if their feature-based similarity and IoU score satisfy $s'_{i, j} > \tau_{sim}$ and $o'_{i,j} > \tau_{iou}$ (same thresholds used during per-frame superpoint merging). Our approach, incorporating matching scores based on point cloud deep features and geometric structures, results in more coherent and well-defined point cloud regions compared to other strategies (see Tab.~\ref{tab:ablation_2d_mask_guided}).





\myheading{Agglomerative clustering process.}
To merge region sets $\{ \r_i\}_{i=1}^I$ and $\{ \r_j\}_{j=1}^J$ from different frames into a unified set $\{ \r_l\}_{l=1}^L$, where $L \leq I+J$, we employ Agglomerative clustering \cite{mullner2011modern}. We begin by concatenating them into a single ``active set'' $\{ \r_l\}_{l=1}^{I+J}$. We compute the each entry $c_{i, j}$ of the binary cost matrix $\C$ of size $(I+J) \times (I+J)$ as:
\begin{equation}
    c_{i,j} = \mathbbm{1} \left ( o'_{i,j} > \tau_{iou} \right ) \odot \mathbbm{1} \left ( s'_{i,j}> \tau_{sim} \right ),
\end{equation}
where $\mathbbm{1}(\cdot)$ is the indicator function, $\odot$ is the AND operator.
%
%
The agglomerative clustering procedure iteratively merges regions within the ``active set'' according to the cost matrix $\C$ and continues to update this matrix until no further merges are possible - indicated by the absence of any positive elements in $\C$.




\myheading{Merging order.} We explored two merging strategies: a \emph{sequential} order, where region sets are merged between consecutive frames, and the resulting set is further merged with the next frame, and a \emph{hierarchical} order, which involves merging region sets between non-consecutive frames in separate passes. The hierarchical approach forms a binary tree, with each level merging sets from consecutive pairs of the previous level (see Fig.~\ref{fig:2dmaskguide}). Details and performance analysis are presented in the Experiments section.

\subsection{3D Instance Segmentation Network} 

\myheading{Network design.} This network directly processes 3D point clouds to generate 3D object instance masks. We employ established 3D instance segmentation networks like Mask3D \cite{Schult23ICRA} and ISBNet \cite{ngo2023isbnet} as our backbone. For each object candidate, the kernel computed from sampled points and their neighbors is convolved with point-wise features to predict the binary mask. In our open-vocabulary scenario, we exclude semantic labeling heads, focusing solely on the binary instance mask head. The output consists of $K_2$ binary masks in a $K_2 {\times} N$ binary matrix $\M_2$ (see Fig.~\ref{fig:architecture} - \textbf{\Circled[]{4}}). 


\begin{figure}[t]
  \centering
  \includegraphics[width=0.98\linewidth]{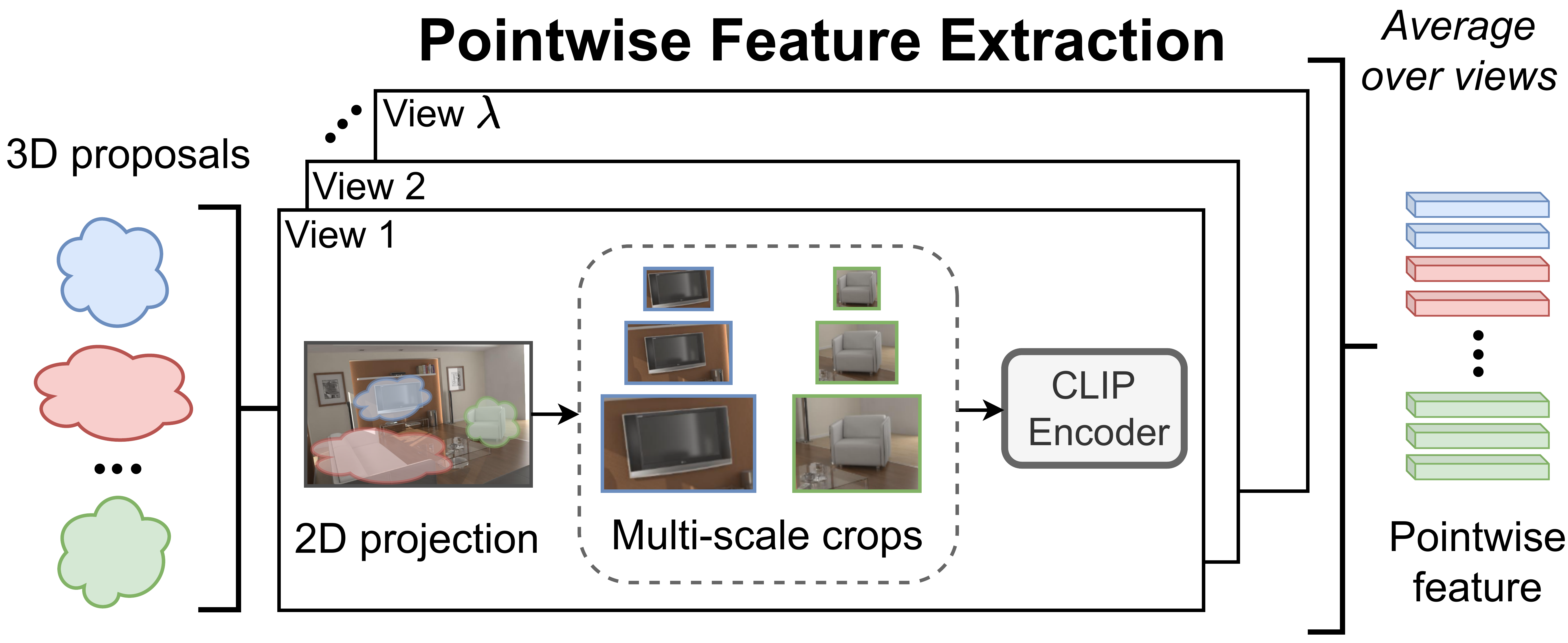}
   \vspace{-8pt}
   \caption{\textbf{Pointwise Feature Extraction}. Each 3D proposal undergoes projection onto top-$\lambda$ views and multiscale cropping \cite{openmask3d}, to extract CLIP features. The resulting proposal feature is then averaged across views and accumulated into the point cloud feature. 
   }
   \label{fig:featurerefinement}
   \vspace{-14pt}
\end{figure}

\myheading{Combining object instance proposals.} We simply append the proposals of set $\M_2$ to $\M_1$ to form the final set of $K$ proposals $\M$ with the size of $K {\times} N$. Note that we apply NMS here to remove near-duplicate proposals with the overlapping IoU threshold $\tau_{dup}$.

\subsection{Pointwise Feature Extraction}
\label{subsec:PFE}
In the final stage of our pipeline, we compute a feature vector for each 3D object proposal from our combined proposal set. This per-proposal feature vector serves various instance-based tasks, such as comparison with text prompts in the CLIP space \cite{radford2021clip}. Unlike prior open-vocabulary instance segmentation methods \cite{openmask3d}, which use a top-$\lambda$ frame/view approach, we employ a more ``3D-aware'' pooling strategy. This strategy accumulates feature vectors on the point cloud, considering the frequency of each point's visibility in each view (see Fig.~\ref{fig:featurerefinement}). \textit{Our rationale is that points more frequently visible in the top-$\lambda$ views should contribute more to the proposal's feature vector}. 


Let $\f^{\text{CLIP}}_{\lambda, k} \in \mathbb{R}^{D^{\text{CLIP}}}$ be the 2D CLIP image feature of $k$-th instance in $\lambda$-th view, $\nu_{\lambda} \in \{0, 1\}^{N}$ be the visibility map of view $\lambda$, and $\m^{\text{3D}}_k \in \{0, 1\}^{N}$ be the $k$-th proposal binary mask in $\M$. We obtain the pointwise CLIP feature $\F^{\text{CLIP}} \in \mathbb{R}^{N \times D^{\text{CLIP}}}$ as:
\begin{align}
    \F^{\text{CLIP}} = \text{NV} \left( \sum_{k} \left(\sum_{\lambda} (\nu_{\lambda} * \f^{\text{CLIP}}_{\lambda, k})* \m^{\text{3D}}_k \right) \right), 
\end{align}
where $*$ is the element-wise multiplication (broadcasting if necessary) and $\text{NV}(x)$ is the L2 normalized vector of $x$.

The final score between a text query $\rho$ and a 3D mask $\m^{\text{3D}}_k$ is the average cosine similarity between its CLIP text embedding $\e_{\rho}$ and all points within the mask, particularly:
\begin{equation}
    s^{\text{CLIP}}_{k, \rho} = \frac{1}{|\m^{\text{3D}}_{k}|} \sum_n  \text{cos}(\F^{\text{CLIP}} * \m^{\text{3D}}_{k}, \e_\rho),
\end{equation}
where $|\m^{\text{3D}}_{k}|$ is the number of points in the $k$-th mask.
\section{Experiments}
\label{sec:exp}

\subsection{Experimental Setup}

\myheading{Datasets.} We mainly conduct our experiments on the challenging dataset ScanNet200 \cite{scannet200}, comprising 1,201 training and 312 validation scenes with 198 object categories. This dataset is well-suited for evaluating real-world open-vocabulary scenarios with a long-tail distribution. Additionally, we conduct experiments on Replica \cite{straub2019replica} (48 classes) and S3DIS \cite{armeni2017joints3dis} (13 classes) for comparison with prior methods \cite{pla, ding2023lowis3d}. Replica has 8 evaluation scenes, while S3DIS includes 271 scenes across 6 areas, with Area 5 used for evaluation. 
We follow the categorization approach from \cite{pla} for S3DIS. Notably, we omit experiments on ScanNetV2 \cite{dai2017scannet} due to its relative ease compared to ScanNet200 and identical input point clouds.

\myheading{Evaluation metrics.} 
We evaluate using standard AP metrics at IoU thresholds of 50\% and 25\%. Additionally, we calculate mAP across IoU thresholds from 50\% to 95\% in 5\% increments. For ScanNet200, we report category group-specific AP$_{\text{head}}$, AP$_{\text{com}}$, and AP$_{\text{tail}}$.

\myheading{Implementation Details.} To process ScanNet200 and S3DIS scans efficiently, we downsampled the RGB-D frames by a factor of 10. Our approach utilizes the Grounded-SAM framework\footnote{\url{https://github.com/IDEA-Research/Grounded-Segment-Anything}}. We employ the dataset class names as text prompts for generating 2D instance masks, followed by NMS with $\tau_{dup}=0.5$ to handle overlapping instances. Our implementation of generating superpoints is from \cite{landrieu2019point,robert2023efficient}. In Pointwise Feature Extraction, each proposal is projected into all viewpoints, and we select the top $\lambda{=}5$ views with the largest number of projected points. For CLIP, we use the \textit{ViT-L/14} \cite{radford2021clip}. We follow OpenMask3D\cite{openmask3d} by setting the confidence score at $1.0$ for every 3D proposal.

\begin{table*}
\small
\setlength{\tabcolsep}{6pt}
\centering
\begin{tabular}{lcccccccc}
\toprule
\textbf{Method} & \textbf{Setting} & \textbf{3D Proposal} & \textbf{AP} & \textbf{AP$_{50}$} & \textbf{AP$_{25}$}  & \textbf{AP}$_{\text{head}}$ & \textbf{AP}$_{\text{com}}$ & \textbf{AP}$_{\text{tail}}$ \\ 
\midrule

\rowcolor{gray!30} ISBNet \cite{ngo2023isbnet}  & & & 24.5 & 32.7 & 37.6  & 38.6 & 20.5 & 12.5  \\
\rowcolor{gray!30} Mask3D \cite{Schult23ICRA} &   \multirow{-2}{*}{Fully-sup} & & 26.9 & 36.2 & 41.4 & 39.8 & 21.7 & 17.9  \\
\midrule
OpenScene \cite{openscene} + DBScan \cite{ester1996density}$^{\dagger}$ & & None & 2.8 & 7.8 & 18.6 & 2.7 & 3.1 & 2.6 \\
OpenScene \cite{openscene} + Mask3D \cite{Schult23ICRA} & & Mask3D \cite{Schult23ICRA} & 11.7 & 15.2 & 17.8 & 13.4 & 11.6 & 9.9\\
SAM3D$^{\dagger}$ \cite{sam3d} & & None &  6.1 & 14.2 & 21.3 & 7.0 & 6.2 & 4.6\\
OVIR-3D$^{\dagger}$ \cite{ovir3d} & & None &  13.0 & 24.9 & \underline{32.3} & 14.4 & 12.7 & 11.7 \\
OpenIns3D \cite{huang2023openins3d} &  & Mask3D \cite{Schult23ICRA}& 8.8 & 10.3 & 14.4 & 16.0 & 6.5 & 4.2 \\
OpenMask3D \cite{openmask3d} & \multirow{-5}{*}{Open-vocab} & Mask3D \cite{Schult23ICRA} & 15.4 & 19.9 & 23.1 & 17.1 & 14.1 & 14.9 \\
\midrule
\textbf{Ours} (only 2D) & & None & 18.2 & \underline{26.1} & 31.4 & 18.9 & 16.5 & \underline{19.2} \\
\textbf{Ours} (only 3D) & & ISBNet \cite{ngo2023isbnet} & \underline{18.6} & 23.1 & 27.3 & \underline{24.7} & \underline{16.9} & 13.3 \\
\textbf{Ours} (2D and 3D) & \multirow{-3}{*}{Open-vocab} & ISBNet \cite{ngo2023isbnet} &  \textbf{23.7 }& \textbf{29.4} & \textbf{32.8} & \textbf{27.8} & \textbf{21.2} & \textbf{21.8} \\

\bottomrule
\end{tabular}
\vspace{-4pt}
\caption{OV-3DIS results on \textbf{ScanNet200}. Methods with $^{\dagger}$ are adapted and evaluated on ScanNet200. Our proposed method achieves
the highest AP, outperforming previous methods in all metrics. The best results are in \textbf{bold} while the second best results are \underline{underscored}.}
\label{tab:scannet200_quanti}
\vspace{-10pt}
\end{table*}
\begin{table}
\small
\setlength{\tabcolsep}{4pt}
\centering
\begin{tabular}{lcccc}
\toprule
\textbf{Method}  & \textbf{Pretrain} & \textbf{AP}$_{\text{novel}}$ & \textbf{AP}$_{\text{base}}$ & \textbf{AP} \\ 
\midrule
\rowcolor{gray!30} OpenMask3D &  & 15.0 & 16.2 & 15.4 \\ 
\rowcolor{gray!30} \textbf{Ours} & \multirow{-2}{*}{ScanNet200} & 22.6 & 26.7 & 23.7 \\ 
\midrule
PLA (Base 15) &  & 0.3 & 10.8 & 3.2 \\ 
PLA (Base 20) &  & 0.3 & 15.8 & 4.5 \\ 
OpenScene + Mask3D & & 7.6 & 11.1 & 8.5 \\ 
OpenMask3D & & 11.9 & 14.3 & 12.6 \\ 
\textbf{Ours} & \multirow{ -5}{*}{ScanNet20} & \textbf{16.5} & \textbf{25.8} & \textbf{19.0} \\ 
\bottomrule
\end{tabular}
\vspace{-4pt}
\caption{OV-3DIS results on \textbf{ScanNet200} dataset, using the class-agnostic 3D proposal network trained on ScanNet20.}
\label{tab:scannet200_pretrain20_quanti}    
\vspace{-5pt}
\end{table}

\begin{figure*}[t]
  \centering
\includegraphics[width=1.\linewidth]{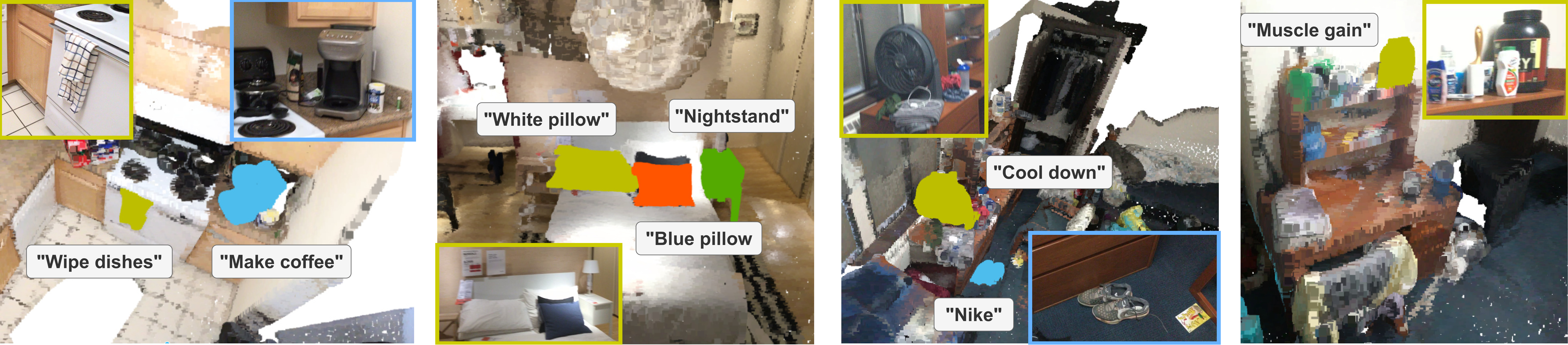}
   \caption{Qualitative results of our method on open-vocabulary instance segmentation. We query instance masks using arbitrary text prompts involving object categories that are not present in the ScanNet200 labels. For each scene, we showcase the instance that has the highest similarity score to the query's embedding. These visualizations underscore the model's open-vocabulary capability, as it successfully identifies and segments objects that were never encountered during the training phase of the 3D proposal network.}
   \label{fig:qualitative}
   \vspace{-10pt}
\end{figure*}

\begin{table}
\small
\setlength{\tabcolsep}{4pt}
\centering
\begin{tabular}{lcccc}
\toprule
\textbf{Method} & \textbf{3D Proposal} & \textbf{AP} & \textbf{AP$_{50}$} & \textbf{AP$_{25}$}\\ 
\midrule
OpenScene + Mask3D  & Mask3D & 10.9 & 15.6 & 17.3 \\ 
OpenMask3D  & Mask3D & 13.1 & 18.4 & 24.2 \\ 
OVIR-3D $^\dagger$  & None & 11.1 & 20.5 & 27.5 \\ 
\midrule
\textbf{Ours} (only 2D) & None & 18.1 & \textbf{26.7} & \textbf{30.5} \\ 
\textbf{Ours} (only 3D) & ISBNet & 14.9 & 18.8 & 23.6 \\ 
\textbf{Ours} (2D and 3D) & ISBNet & \textbf{18.5} & 24.5 & 28.2 \\ 
\bottomrule
\end{tabular}
\vspace{-4pt}
\caption{OV-3DIS results on \textbf{Replica} dataset.${^\dagger}$We adopt the source code of \cite{ovir3d} to this dataset.}
\label{tab:replica}    
    \vspace{-12pt}
\end{table}



\begin{table}
\small
\setlength{\tabcolsep}{8pt}
\centering
\begin{tabular}{lcccc}
\toprule
\multirow{ 2}{*}{\textbf{Method}} & \multicolumn{2}{c}{\textbf{B8/N4}} & \multicolumn{2}{c}{\textbf{B6/N6}} \\
 & \textbf{AP}$^{B}_{50}$ & \textbf{AP}$^N_{50}$ & \textbf{AP}$^{B}_{50}$ & \textbf{AP}$^N_{50}$ \\
\midrule
LSeg-3D \cite{pla} & 58.3 & 0.3 & 41.1 & 0.5 \\
PLA \cite{pla} & 59.0 & 8.6 & 46.9 & 9.8 \\
Lowis3D \cite{ding2023lowis3d} & 58.7 & 13.8 & \textbf{51.8} & 15.8 \\
\midrule
\textbf{Ours} & \textbf{60.8} & \textbf{26.3} & 50.0 & \textbf{29.0} \\
\bottomrule
\end{tabular}
\caption{OV-3DIS results on S3DIS in terms of  \textbf{AP}$^{B}_{50}$ and \textbf{AP}$^N_{50}$.}
\label{tab:s3dis_quanti}
\vspace{-8pt}
\end{table}

\subsection{Comparison to prior work}
\myheading{Setting 1: ScanNet200.} 
The quantitative evaluation of the ScanNet200 dataset is summarized in Tab.~\ref{tab:scannet200_quanti}.
Following \cite{openmask3d}, we utilize the class-agnostic 3D proposal network trained on the ScanNet200 training set, then test the OV-3DIS on the validation set.  
Employing our 2D-Guided-3D Instance Proposal Module, \Approach~ achieves $18.2$ and $19.2$ in AP and AP$_{\text{tail}}$. We outperform OVIR-3D \cite{ovir3d} and OpenMask3D \cite{openmask3d} by margins of $+5.2$ and $+2.8$ in AP, and surpass all other methods, even the fully-supervised approaches in the AP$_{\text{tail}}$ metric. This emphasizes the effectiveness of our 2D-Guided-3D Instance Proposal Module, which is effective in crafting precise 3D instance masks independently of any 3D models. Combining with class-agnostic 3D proposals from ISBNet boosts our performance to $23.7$, $29.4$, and $32.8$ in AP, AP$_{50}$, and AP$_{25}$ — reflecting a $1.5\x$ enhancement in AP compared to prior methods. 
Impressively, our method competes closely with fully supervised techniques, attaining approximately $96\%$ and $88\%$ of the AP scores of ISBNet and Mask3D, and excelling in the AP$_{\text{com}}$ and AP$_{\text{tail}}$. This performance underscores the advantages of merging 2D and 3D proposals and demonstrates our model's adeptness at segmenting rare objects. 

\begin{table}
    \small
    \setlength{\tabcolsep}{3pt}
    \centering
    \begin{tabular}{lcccc}
    \toprule
    \textbf{Setting} & \textbf{AP}  & \textbf{AP}$_{\text{head}}$ & \textbf{AP}$_{\text{com}}$ & \textbf{AP}$_{\text{tail}}$ \\
    \midrule
    A1: OpenScene (distill) & 3.3 & 5.5 & 2.4 & 1.7 \\
    A2: OpenScene (fusion) & 17.5 & 21.5 & 17.1 & 13.3 \\
    A3: OpenScene (ensemble) & 5.6 & 6.4 & 4.8 & 5.7 \\
    B: Mask-wise Feature & 22.2 & 25.9 & 19.3 & 21.4 \\
    C: Point-wise Feature & \textbf{23.7} & \textbf{27.8} & \textbf{21.2} & \textbf{21.8} \\
    \bottomrule
    \end{tabular}  
    \vspace{-5pt}    
    \caption{Comparing between extracting per-mask and per-point features for classification using \Approach~ instance proposal set.} 
    \label{tab:ablation_classification_feat}
    \vspace{-10pt}
\end{table}

\begin{table}
\small
\setlength{\tabcolsep}{3pt}
\centering
\begin{tabular}{cccccc}
\toprule
\textbf{Use Superpoint} & \textbf{Filtering Cond.} & \textbf{AP} & \textbf{AP}$_{\text{head}}$ & \textbf{AP}$_{\text{com}}$ & \textbf{AP}$_{\text{tail}}$\\
\midrule
 \checkmark &  Deep. Feature & \textbf{18.2} & \textbf{18.9} &  \textbf{16.5} & \textbf{19.2} \\ 
\checkmark & None  & 15.9 & 16.5  & 14.3 & 17.0 \\
\checkmark & Euclid Dist. & 16.0 & 16.4  & 14.1 & 17.6 \\
 & None   & 12.0 & 12.6  & 11.2 & 12.2 \\
\bottomrule
\end{tabular}
\vspace{-5pt}
\caption{Ablation on different configurations of the 2D-G-3DIP.}
\vspace{-7pt}
\label{tab:ablation_2d_mask_guided}
\end{table}

To assess the generalizability of our approach, we conducted an additional experiment where the class-agnostic 3D proposal network is substituted with the one trained solely on the ScanNet20 dataset. We then categorized the ScanNet200 instance classes into two groups: the \textit{base} group, consisting of 51 classes with semantics similar to ScanNet20 categories, and the \textit{novel} group of the remaining classes. We report the AP$_{\text{novel}}$, AP$_{\text{base}}$, and AP in Tab.~\ref{tab:scannet200_pretrain20_quanti}. Our proposed ~\Approach~ achieves superior performance compared to PLA \cite{pla}, OpenMask3D \cite{openmask3d}, with large margins in both \textit{novel} and \textit{base} classes. Notably, PLA \cite{pla}, trained with contrastive learning techniques, falls in a setting with hundreds of novel categories. 

\myheading{Setting 2: Replica}. We further evaluate the zero-shot capability of our method on the Replica dataset, with results detailed in Tab.~\ref{tab:replica}. Considering that several Replica categories share semantic similarities with ScanNet200 classes, to maintain a truly zero-shot scenario, we omitted the class-agnostic 3D proposal network for this dataset (using proposals from 2D only).
Under this constraint, our approach still outperforms OpenMask3D \cite{openmask3d} and OVIR-3D \cite{ovir3d} by margins of $+5.0$ and $+7.0$ in AP, respectively.


\myheading{Setting 3: S3DIS.} In line with the setting of PLA \cite{pla}, we trained a fully-supervised 3DIS model on the \textit{base} classes of the S3DIS dataset, followed by testing the model on both \textit{base} and \textit{novel} classes. The results are shown in Tab.~\ref{tab:s3dis_quanti}, where we report the performance in terms of AP$^{B}_{50}$ and AP$^N_{50}$, representing the AP$_{50}$ for the \textit{base} and \textit{novel} categories, respectively. \Approach~significantly outperforms existing methods in AP$^N_{50}$, achieving more than double their scores. This remarkable performance underscores the efficacy of our approach in dealing with unseen categories, with the support of the 2D foundation model. 

\begin{table}
    \small
    \setlength{\tabcolsep}{3pt}
    \centering
    \begin{tabular}{llcccc}
    \toprule
    \textbf{Merging Strat.} & \textbf{Merging Ord.} &  \textbf{AP}  & \textbf{AP}$_{\text{head}}$ & \textbf{AP}$_{\text{com}}$ & \textbf{AP}$_{\text{tail}}$ \\
    \midrule
    Hungarian & Sequential & 13.2 & 13.9 & 11.3 & 14.7 \\
    Hungarian & Hierarchical & 16.1 & 16.1 & 13.3 & 19.4   \\
    Agglomerative  & Sequential  & 16.9 & 17.8 & 16.1 & 18.0 \\
    Agglomerative  & Hierarchical  & \textbf{18.2} & \textbf{18.9} & \textbf{16.5} & \textbf{19.2}  \\
    \bottomrule
    \end{tabular}      
    \vspace{-5pt}
    \caption{Ablation on different merging configurations.
    } 
    \label{tab:ablation_traversal}
    \vspace{-8pt}
\end{table}

\myheading{Our qualitative results with arbitrary text queries}. We visualize the qualitative results of text-driven 3D instance segmentation in Fig.~\ref{fig:qualitative}. Our model successfully segments instances based on different kinds of input text prompts, involving object categories that are not present in the labels, object's functionality, object's branch, and other properties.


\subsection{Ablation study}
To validate design choices of our method, series of ablation studies are conducted on validation set of ScanNet200.

\myheading{Study on different kinds of features for open-vocabulary classification} is presented in Tab.~\ref{tab:ablation_classification_feat}. In the first three rows (setting A1-A3), we employ the pointwise feature map extracted by OpenScene \cite{openscene} to perform classification on our 3D proposals. Of these, the \textit{fusion} approach, which directly projects CLIP features from 2D images onto the 3D point cloud, yields the highest results, $17.5$ in AP. In setting B, we adopt a strategy akin to \cite{openmask3d}, extracting features for each mask by projecting the 3D proposals onto the top-$\lambda$ views, which attains an AP of $22.2$. Surpassing these, our Pointwise Feature Extraction (setting C) achieves the best AP score of $23.7$, substantiating our design choice.


\myheading{Study on the 2D-Guided-3D Instance Proposal Module} is in Tab. \ref{tab:ablation_2d_mask_guided}. Our proposed approach (row 1), utilizing superpoints to merge 3D points into regions and filter outliers based on cosine similarity in feature space, achieves an AP of 18.2. Disabling this filtering notably reduces AP by 2.3.
Comparatively, a more basic method (row 3) relying on Euclidean distance to eliminate outlier superpoints yields an AP of 16.0, showing the lesser effectiveness of Euclidean distance for noise filtering.
Our baseline (last row), grouping 3D points solely based on 2D masks, significantly decreases AP to 12.0, underscoring the necessity of superpoint merging for effective 3D proposal creation.




We study different merging configurations, including \textit{merging strategy} and \textit{merging order} in Tab. \ref{tab:ablation_traversal}. Specifically, we first establish a partial matching between two sets of regions, then matched pairs are merged into new refined regions, and unmatched ones remain the same. Using Hungarian matching yields inferior results relative to proposed Agglomerative Clustering, with a drop of ${\sim}2.0$ in AP. Adopting the sequential merging order leads to a slight decrease by ${\sim}1.0$ in AP in performance. The best results are achieved when agglomerative clustering is paired with the hierarchical merging order.



\begin{table}
    \small
    \setlength{\tabcolsep}{6pt}
    \centering
    \begin{tabular}{lcccc}
    \toprule
    \textbf{3D Seg.} & \textbf{AP}  & \textbf{AP}$_{\text{head}}$ & \textbf{AP}$_{\text{com}}$ & \textbf{AP}$_{\text{tail}}$ \\
    \midrule
    Mask3D \cite{Schult23ICRA} & \textbf{23.7} & 26.4 & \textbf{22.5} & \textbf{21.9} \\
    ISBNet \cite{ngo2023isbnet} & \textbf{23.7} & \textbf{27.8} & 21.2 & 21.8  \\
    \bottomrule
    \end{tabular}          
    \vspace{-5pt}
    \caption{Ablation on different 3D segmenters.} 
    \label{tab:ablation_3dsegmenter}
    \vspace{-8pt}
\end{table}

\begin{table}
    \small
    \setlength{\tabcolsep}{6pt}
    \centering
    \begin{tabular}{lcccc}
    \toprule
    \textbf{2D Seg.} & \textbf{AP}  & \textbf{AP}$_{\text{head}}$ & \textbf{AP}$_{\text{com}}$ & \textbf{AP}$_{\text{tail}}$ \\
    \midrule
    SEEM \cite{zou2023segment} & 21.5 & 26.5 & 19.6 & 18.0 \\
    ODISE \cite{xu2023odise} & 21.6 & 26.0 & 19.5 & 19.1 \\
    Detic \cite{zhou2022detecting} & 22.2 & 26.8 & 20.0 & 19.2 \\
    Grounded-SAM  & \textbf{23.7} & \textbf{27.8} & \textbf{21.2} & \textbf{21.8} \\
    \bottomrule
    \end{tabular}       
    \vspace{-5pt}   
    \caption{Ablation on different 2D segmenters.} 
    \label{tab:ablation_2dsegmenter}
    \vspace{-8pt}
\end{table}

\myheading{Ablation Study on Segmenters.} Our comparative analysis of various \textit{class-agnostic 3D segmenters} and \textit{open-vocabulary 2D segmenters} is presented in Tab.~\ref{tab:ablation_3dsegmenter} and~\ref{tab:ablation_2dsegmenter}. The findings reveal that utilizing either ISBNet \cite{ngo2023isbnet} or Mask3D \cite{Schult23ICRA} leads to similar levels of performance, achieving an AP of $23.7$. Incorporating 2D instance masks from SEEM \cite{zou2023segment}, Detic \cite{zhou2022detecting} or ODISE \cite{xu2023odise} leads to a slight decrease in AP by ${\sim}1.4$, which we attribute to the less refined outputs produced by these models. 

\myheading{Ablation study on different values of visibility threshold and similarity threshold}. We report the performance of our version using only proposals from the 2D-G-3DIP with different values of the visibility threshold and similarity threshold in Tab.~\ref{tab:ablation_visi} and~\ref{tab:ablation_simi}.

\begin{table}
    \parbox{.45\linewidth}{
        \small
        \setlength{\tabcolsep}{1pt}
        \centering
        \begin{tabular}{lccccc}
        \toprule
        $\tau_{iou}$ & 0.3  & 0.5 & 0.7 & 0.9 & 0.95 \\
        \midrule
        \textbf{AP} & 17.7 & 17.8 & 18.0 & \textbf{18.2} & 16.9\\
        \textbf{AP$_{50}$} & 25.4 & 25.8 & 25.9 & \textbf{26.1} & 24.1 \\
        \bottomrule
        \end{tabular}      
        \vspace{-4pt}
        \caption{Ablation on $\tau_{iou}$.}
        \label{tab:ablation_visi}
    }
    \hfill
    \parbox{.52\linewidth}{
        \small
        \setlength{\tabcolsep}{1pt}
        \centering
        \begin{tabular}{lccccc}
        \toprule
        $\tau_{sim}$ & 0.5 & 0.7 & 0.8  & 0.9 & 0.95 \\
        \midrule
        \textbf{AP}  & 14.2 & 14.6 & 17.2 & \textbf{18.2} & 16.2 \\
        \textbf{AP$_{50}$} & 21.0 & 21.8 & 25.1 & \textbf{26.1} & 23.8 \\
        \bottomrule
        \end{tabular}
        \vspace{-4pt}
        \caption{Ablation on $\tau_{sim}$.}
        \label{tab:ablation_simi}
        }
    \vspace{-7pt}
\end{table}

\begin{table}
    \small
    \setlength{\tabcolsep}{3pt}
    \centering
    \begin{tabular}{lccccc}
    \toprule
    \textbf{View Selection}  & Top 1  & Top 5 & Top 10 & Top 20 & All \\
    \midrule
    \textbf{AP} & 21.2 & \textbf{23.7} & 22.6 & 22.5 & 22.5\\
    \textbf{AP$_{50}$} & 27.3 & \textbf{29.4} & 28.7 & 29.0 & 29.1\\
    \bottomrule
    \end{tabular}      
    \caption{Ablation on top-$\lambda$ view selection.} 
    \label{tab:ablation_topk_view}
    \vspace{-12pt}
\end{table}

\myheading{Study on different values of viewpoints} is illustrated in Tab.~\ref{tab:ablation_topk_view}. Relying only on the viewpoint with the highest number of projected points reduces the AP score to $21.2$. Conversely, raising the number of views to 10 or more also yields worse results, likely due to the presence of inferior, occluded 2D masks. $\lambda{=}5$ reports the best performance.

\section{Discussion}
\label{sec:conclusion}

We presented a method for open-vocabulary instance segmentation in 3D scenes, which aggregates proposals from both point cloud-based instance segmenters and 2D image-based networks in a geometrically coherent manner.

\myheading{Limitations.} Our Class-agnostic 3D Proposal and 2D-Guided-3D Instance Proposal Module currently operate independently, with their outputs being combined to obtain the final 3D proposal set. A better-integrating strategy, where these modules enhance each other's performance in a synergistic fashion, would be an interesting future direction.


{\small

\bibliographystyle{ieeenat_fullname}
\bibliography{main}
}






\def\Approach{Open3DIS}


\clearpage

\section{Implementation Details}
\subsection{Class-agnostic 3D Segmenter}
\label{sec:cls_agnostic_3d_segmenter}
We adopt the architecture from ISBNet \cite{ngo2023isbnet} to serve as our class-agnostic 3D proposal network due to its publicly released implementation. This network processes $N$ points in a colored point cloud
$\P \in \mathbb{R}^{N \times 6}$ and outputs a collection of $K$ binary 3D instance mask $\M \in \{0,1\}^{K \times N}$. At its core is a 3D UNet backbone a 3D UNet backbone \cite{graham20183d}, utilizing 3D sparse convolutions \cite{graham2017submanifold}, which processes the input to produce a feature map $\F^{3D}$ 
of the point cloud. Subsequently, an instance-wise encoder, based on a sampling strategy, refines these features to produce instance-specific kernels and bounding box parameters. The final stage involves a box-aware dynamic convolution, which employs these instance kernels and mask features, augmented by the corresponding box predictions, to compute the binary mask for each instance.

During inference, we utilize the Intersection over Union (IoU) prediction score to filter out lower-quality masks, with a threshold of $0.2$. This score is neutral regarding object classes—during training, the IoU prediction head is trained on the IoU values calculated between the predicted masks and their ground truth counterparts, which are determined by the Bipartite Matching algorithm. Next, we employ superpoints \cite{landrieu2019point, robert2023efficient} to refine the alignment of our proposals with the actual point cloud structure. This step ensures that our segmentation is consistent with the spatial organization of the point cloud. Lastly, we discard any small proposals that have fewer than $50$ points.

\subsection{Open-Vocabulary 2D Segmenter}
\label{sec:open_vocab_2d_segmenter}
In this study, we employ four 2D open-vocabulary instance segmenters: Grounded-SAM\footnote{\url{https://github.com/IDEA-Research/Grounded-Segment-Anything}}, DETIC \cite{zhou2022detecting}, SEEM \cite{zou2023segment}, and ODISE \cite{xu2023odise}. Here is a breakdown of how each of these segmenters is utilized:

\noindent (a) \textit{For Grounded-SAM}, we utilize the Swin-B Grounding DINO decoder \cite{li2023grounded}, which has been pretrained on various datasets including COCO \cite{lin2014microsoft}, O365 \cite{shao2019objects365}, GoldG \cite{krishna2017visual, plummer2015flickr30k}, OpenImage \cite{kuznetsova2020open}, ODinW-35 \cite{li2022elevater}, and RefCOCO \cite{kazemzadeh2014referitgame}. This model is employed to generate bounding boxes from a given text prompt, with box and text thresholds both set to 0.4. Subsequently, these generated bounding boxes are passed through the ViT-L Segment Anything Model \cite{kirillov2023segany} to produce instance masks. To process every text query caption, we divide it into chunks, each containing 10 classes, accommodating the limitations of the 77-token decoder. Finally, we apply Non-Maximum-Suppression with an IoU threshold of 0.5 to obtain the ultimate bounding boxes.

\noindent  (b) \textit{For DETIC}, we follow \cite{ovir3d} to use the Swin-B model pre-trained on the ImageNet-21K dataset \cite{deng2009imagenet} with 21K classes as text queries. We set the confidence threshold at 0.5.

\noindent (c) \textit{For SEEM}, we employ the Focal-T visual decoder, which is trained on RefCOCO and LVIS \cite{gupta2019lvis}, with a logit score threshold of 0.4. Similar to Grounded-SAM, SEEM follows a query processing and post-processing procedure.

\noindent (d) \textit{For ODISE}, we utilize the pre-trained label COCO version. This model is complemented by the Stable Diffusion \cite{rombach2022high} pre-trained on a subset of the LAION \cite{schuhmann2022laion} dataset, along with Mask2Former \cite{cheng2022masked} serving as the mask generator. We set the confidence threshold to 0.5.

\subsection{S3DIS and Replica Datasets}
\noindent (a) \textit{For the S3DIS dataset}, which lacks original mesh data, we apply the superpoint-graph method from the Superpoint Transformer \cite{robert2023spt} to generate superpoints straight from the 3D point cloud data. For scenes having an extra large number of points (e.g. 1M points), we subsample the point cloud by a factor of 4 for efficient processing.

\noindent (b)  \textit{For the Replica dataset}, we adopt the mesh segmentation tool\footnote{\url{https://github.com/ScanNet/ScanNet/tree/master/Segmentator}} based on Felzenszwalb and Huttenlocher's efficient graph-based image segmentation method \cite{felzenszwalb2004efficient} to create superpoints. The ground-truths for semantic and instance segmentation are provided by \cite{openmask3d}.




\subsection{3D Object Proposal Formation Process} \label{sec:3d_formation}
\begin{figure*}[t]
  \centering
\includegraphics[width=0.9\linewidth, height=0.75\paperheight]{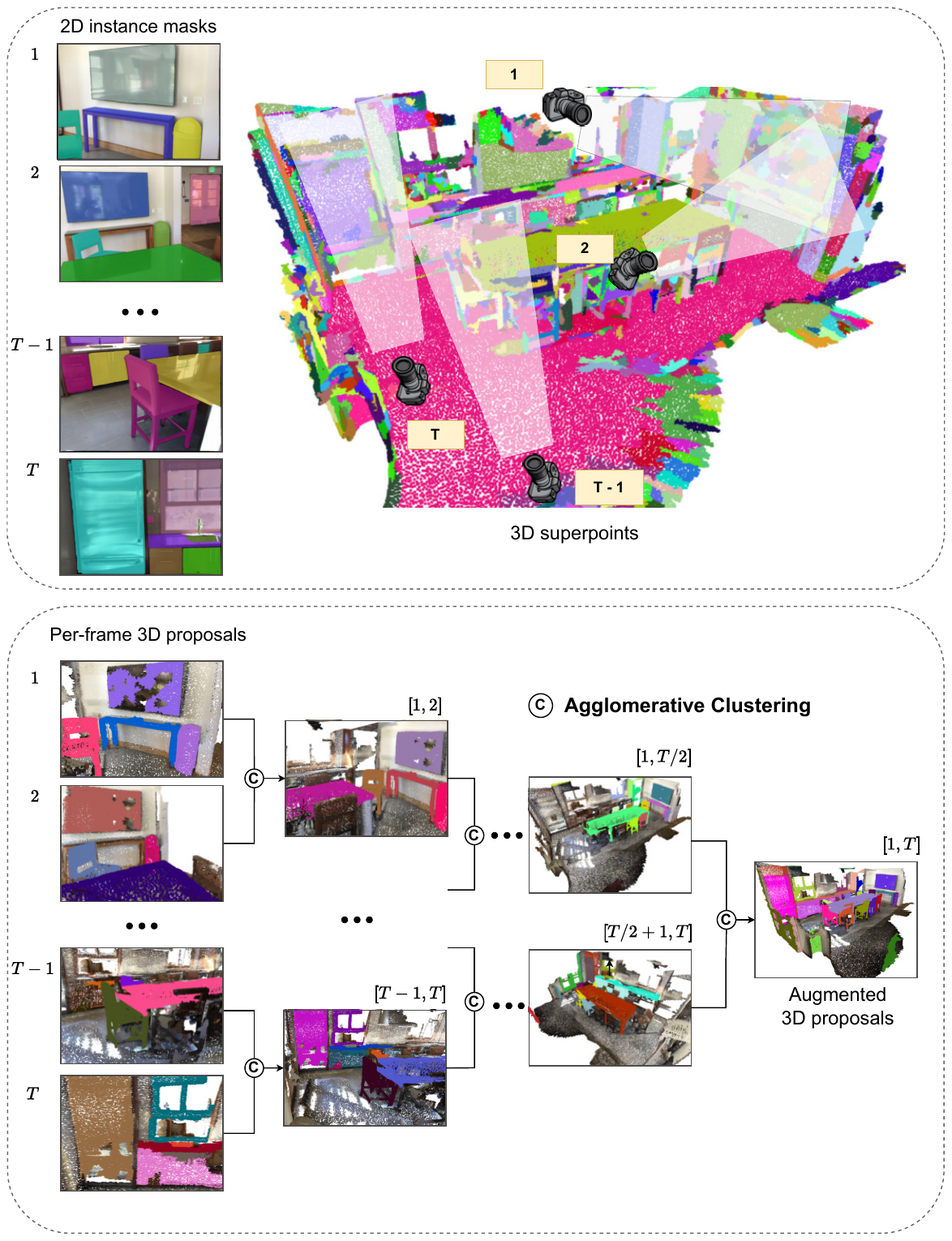}
   \caption{(Top) The 2D-G-3DIP module utilizes 2D per-frame instance masks to generate per-frame 3D proposals by leveraging 3D superpoints. (Bottom) Our proposed hierarchical merging. These proposals are considered point cloud regions and undergo a hierarchical merging process across multiple views, resulting in the final Augmented 3D proposals (Best viewed in color).}
   \label{fig:quali_2d3d}
\end{figure*}

The implementation details of the 3D Object Proposal Formation Process using the \textit{Hierarchical merging order} and \textit{Agglomerative merging strategy}  are shown in Alg.~\ref{alg:obj_proposal_algo}. Having the 3D point cloud regions obtained from the merging procedure across individual frames $\{\r_1, \r_2, \ldots, \r_T\}$, the algorithm merges these independently fragmented regions (see Fig.~\ref{fig:quali_2d3d}) into well-formed ones recursively, resulting in high-quality augmented 3D proposals.


\subsection{Point cloud - Image Projection}
\label{sec:cam_calib}
To establish the correspondence between a 3D point cloud and each frame of the RGB-D sequence $\V$, we employ the principles of pinhole camera projection. Given a 3D point cloud $\P = \{\p_{i}\}_{i=1}^{N} \in \mathbb{R}^{N \times 6}$, and for a specific frame $t$, we consider its depth image $\D_{t} \in \mathbb{R}^{H \times W}$, intrinsic matrix $K_{t} \in \mathbb{R}^{3 \times 3}$ and extrinsic matrix $\mathbf{[R | c]}_{t} \in \mathbb{R}^{3 \times 4}$, where $\R$ is a 3D rotation matrix and $\c$ is a 3D translation vector. The composite matrix of rotation and translation
converts coordinate from the global frame (of the point cloud) to the camera's frame at time $t$. We compute the projection matrix that maps 3D points to 2D image coordinates as follows:

\begin{align}
    \Pi_{t} = \K_{t} \cdot \mathbf{[R | c]}_{t}
\end{align}

Then the 2D projection of a 3D point $\p_{i} = [x^{(3d)}_{i}, y^{(3d)}_{i}, z^{(3d)}_{i}] \in \P$ is given by:

\begin{align}
    z^{(2d)}_{i} \cdot \begin{bmatrix}
x^{(2d)}_{i}\\ 
y^{(2d)}_{i} \\
1
\end{bmatrix} = \Pi_{t} \cdot \begin{bmatrix}
x^{(3d)}_{i}\\ 
y^{(3d)}_{i}\\ 
z^{(3d)}_{i}\\ 
1
\end{bmatrix}
\end{align}
where $z^{(2d)}_{i}$ is the projected depth value and $x^{(2d)}_{i}, y^{(2d)}_{i}$ is the 2D pixel coordinate. Next, we discard any points whose projections fall outside the image boundaries, defined by $x^{(2d)}_{i} \notin [0, W-1]$ or $y^{(2d)}_{i} \notin [0, H-1]$. To address occlusion within that viewpoint, we further filter out points where the difference between their projected depth and the actual depth recorded at the corresponding pixel in the depth image exceeds a certain depth threshold $\tau_{\text{depth}}$:

\begin{align}
    |z^{(2d)}_{i} - \D_{t}[\lfloor y^{(2d)}_{i} \rfloor, \lfloor x^{(2d)}_{i} \rfloor]| > \tau_{depth}
\end{align}



\begin{algorithm}[tb]
\caption{3D Object Proposal Formation}
\label{alg:obj_proposal_algo}
\textbf{Input: }{$T$ per-frame merged point cloud regions $\{\r_t\}_{t=1}^T$.} \\
\textbf{Output: }{Augmented 3D proposal set $\r$}. 

\begin{algorithmic}[1]

\Function{Hierarchical\_Traverse}{$s$: start, $e$: end}

\If {$s = e$}
    \State \Return $\r_{s}$ \Comment{Look up in $\{\r_t\}_{t=1}^T$}
\Else
    \State $m \gets \left \lfloor (s+e) / 2 \right \rfloor$
    \State $\r_{\text{left}} \gets\Call{Hierarchical\_Traverse}{s, m}$
    \State $\r_{\text{right}} \gets \Call{Hierarchical\_Traverse}{m+1, e}$
    \State $\r \gets (\r_{\text{left}} \cup \r_{\text{right}})$
    \State $\C_{\r} \gets \Call{Cost\_Matrix}{\r}$ \Comment{following Eq. (1) in the main paper}
    \State $\r \gets \Call{Agglomerative\_Clustering}{\r, \C_{\r}}$
    \State \Return $\r$
    
\EndIf
\EndFunction

\State $\r \gets \Call{Hierarchical\_Traverse}{1,T}$
\end{algorithmic}
\end{algorithm}

\section{Additional Analysis}

\myheading{Ablation study on the depth threshold} $\tau_{depth}$ is reported in Tab.~\ref{tab:ablation_depth_thresh}. Overall, $\tau_{depth} = 0.1$ gives the best performance.

\begin{table}
    \small
    \centering
    \begin{tabular}{ccccc}
    \toprule
    $\tau_{\text{depth}}$ &  \textbf{AP}  & \textbf{AP}$_{\text{head}}$ & \textbf{AP}$_{\text{com}}$ & \textbf{AP}$_{\text{tail}}$ \\
    \midrule
    0.2 & 17.4 & 17.7 & 15.6 & 19.3 \\
    0.1 & 18.2 & \textbf{18.9} & 16.5 & 19.2  \\
    0.05 & \textbf{18.7} & 17.7 & 16.4 & \textbf{22.8} \\
    0.025 & 17.7 & 17.6 & \textbf{17.6} & 18.6 \\
    0.01 & 16.7 & 16.3 & 13.8 & 21.2 \\
    \bottomrule
    \end{tabular}      
    \caption{Ablation on the depth threshold $\tau_{\text{depth}}$.} 
    \label{tab:ablation_depth_thresh}
\end{table}


\myheading{Ablation study on the subsampling factors of RGB-D images} is shown in Tab.~\ref{tab:ablation_traversal}. By default, we subsample the number of images by a factor of 10. Increasing the subsampling factor to 20 or 40 slightly decreases the performance to 17.1 in AP scores. Reducing the number of images too much yields worse results. We also report the total runtime (in \textit{hours}) to inference on the whole validation set of ScanNet200 in the last column.

\begin{table}
    \small
    \setlength{\tabcolsep}{3pt}
    \centering
    \begin{tabular}{ccccccc}
    \toprule
    \textbf{Use 3D} & \textbf{Sub. factor} &  \textbf{AP}  & \textbf{AP}$_{\text{head}}$ & \textbf{AP}$_{\text{com}}$ & \textbf{AP}$_{\text{tail}}$ & \textbf{Time} (h)\\
    \midrule
    \checkmark & 10 (default)  & \textbf{23.7} & \textbf{27.8} & 21.2 & 21.8 & 20 + 2.3 \\
    & 10 (default)  & \textbf{18.2} & \textbf{18.9} & 16.5 & 19.2 & 20 + 2 \\
    & 20  & 17.9 & 17.9 & 16.5 & \textbf{19.6} & 10 + 1  \\
    & 40   & 17.4 & 17.3 & \textbf{16.7} & 18.5 & 5 + 0.5 \\
    & 80   & 16.5 & 16.7 & 15.4 & 17.1 \\
    & 160 & 13.2 & 12.4 & 12.4 & 15.2 \\
    & 320 & 9.0 & 8.6 & 8.0 & 10.7 \\
    \bottomrule
    \end{tabular}      
    \caption{Study on the subsampling factors of RGB-D images.} 
    \label{tab:ablation_traversal}
    \vspace{-8pt}
\end{table}

\myheading{Class-agnostic evaluation on ScanNet200 \cite{scannet200} and ScanNet++ \cite{yeshwanthliu2023scannetpp}} We further examine the quality of mask proposals generated by Open3DIS on the ScanNet200 and ScanNet++ datasets. In ScanNet200, employing the 3D backbone ISBNet, Open3DIS (2D + 3D) demonstrates superior performance over existing methods in producing high-quality 3D proposals, as depicted in Tab. \ref{tab:clsagnostic200}. In ScanNet++, unlike previous methods, we utilize only 100 subsampled 2D RGB-D frames per 3D scene (for computational efficiency). The results using solely 2D data exhibit promising outcomes, as illustrated in Tab. \ref{tab:clsagnosticpp}.

\begin{table*}
    \centering
    \begin{tabular}{lcccccc}
    \toprule
    \textbf{Method} & \textbf{AP} & \textbf{AP$_{50}$} & \textbf{AP$_{25}$}  & \textbf{AR} & \textbf{AR$_{50}$} & \textbf{AR$_{25}$}  \\
    \midrule
    \text{Superpoint} & 5.0 & 12.7 & 38.9 \\
    \text{DBSCAN \cite{ester1996density}} & 1.6 & 5.5 & 32.1 \\
    \text{OVIR-3D \cite{ovir3d} (Detic)} & 14.4 & 27.5 & 38.8 \\
    \text{Mask Clustering \cite{yan2024maskclustering} (CropFormer)} & 17.4 & 33.3 & 46.7 \\
    \text{ISBNet \cite{ngo2023isbnet} (3D)} & 40.2 & 50.0 & 54.6 & 66.8 & 80.4 & 87.4 \\
    \hline
    \textbf{Ours} \text{ (Grounded SAM)} & 29.7 & 45.2	& 56.8 &	49.0 & 70.0	 & 83.2 \\
    \textbf{Ours} \text{ (3D + Grounded SAM)} & 34.6 & 43.1 & 48.5 & 66.2 & 81.6 & 91.4 \\
    \textbf{Ours} \text{ (SAM)} & 31.5 & 45.3	& 51.1 & 61.2 & 87.1 & 97.5 \\
    \textbf{Ours} \text{ (3D + SAM)} & 41.5 & 51.6 & 56.3 & 74.8 & 90.9 & 97.8 \\
    \bottomrule
    \end{tabular}      
    \caption{Class-agnostic evaluation on ScanNet200 \cite{scannet200} (updated on 2024, Mar. 19th).}
    \label{tab:clsagnostic200}
\end{table*}

\begin{table*}
    \centering
    \begin{tabular}{lccccccc}
    \toprule
    \textbf{Method} & \textbf{AP} & \textbf{AP$_{50}$} & \textbf{AP$_{25}$}  & \textbf{AR} & \textbf{AR$_{50}$} & \textbf{AR$_{25}$}  & \textbf{NOTE} \\
    \midrule
    \text{ISBNet \cite{ngo2023isbnet} (3D)} & 6.2 & 10.1  &	16.2  & 10.9 & 16.9 & 25.2 & \text{pretrained Scannet200}\\
    \hline
    \text{SAM3D \cite{sam3d}} & 7.2 & 14.2 & 29.4 &\\
    \text{SAM-guided Graph Cut \cite{guo2023sam-graph}} & 12.9 & 25.3 & 43.6 &\\
    \text{Segment3D \cite{Huang2023Segment3D}} & 12.0 & 22.7 & 37.8 &\\
    \text{SAI3D \cite{yin2023sai3d} (SAM)}  & 17.1 & 31.1 & 49.5 &\\
    \hline
    \textbf{Ours} \text{(SAM)} & 18.5 & 33.5	& 44.3 &	35.6 & 63.7	 & 82.7 &  \text{100 frames per scene} \\
    \textbf{Ours} \text{(SAM)} & 20.7 & 38.6	& 47.1 &	40.8 & 75.7	 & 91.8 &  \text{all frames per scene}\\
    \bottomrule
    \end{tabular}      
    \caption{Class-agnostic evaluation on ScanNet++ \cite{yeshwanthliu2023scannetpp} (updated on 2024, Mar. 19th).}
    \label{tab:clsagnosticpp}
\end{table*}

To assess the quality of class-agnostic masks in the 2D context, we utilize all masks generated by the 2D-G-3DIP module without any postprocessing, which typically yields high recall albeit at the cost of precision. In the case of 3D masks, we select the top 100 masks from ISBNet based on their confidence scores. Subsequently, to evaluate the Open-Vocab capability, the class-agnostic masks undergo postprocessing by selecting the top k (where k ranges approximately between 300 and 600) masks with the highest CLIP scores. Final confidence score set to 1.0 (OpenMask3D).

\section{Qualitative Results}

\subsection{Constructing 3D proposals from a single image}
\label{sec:single_img}
\begin{figure*}[t]
  \centering
\includegraphics[width=1\linewidth]{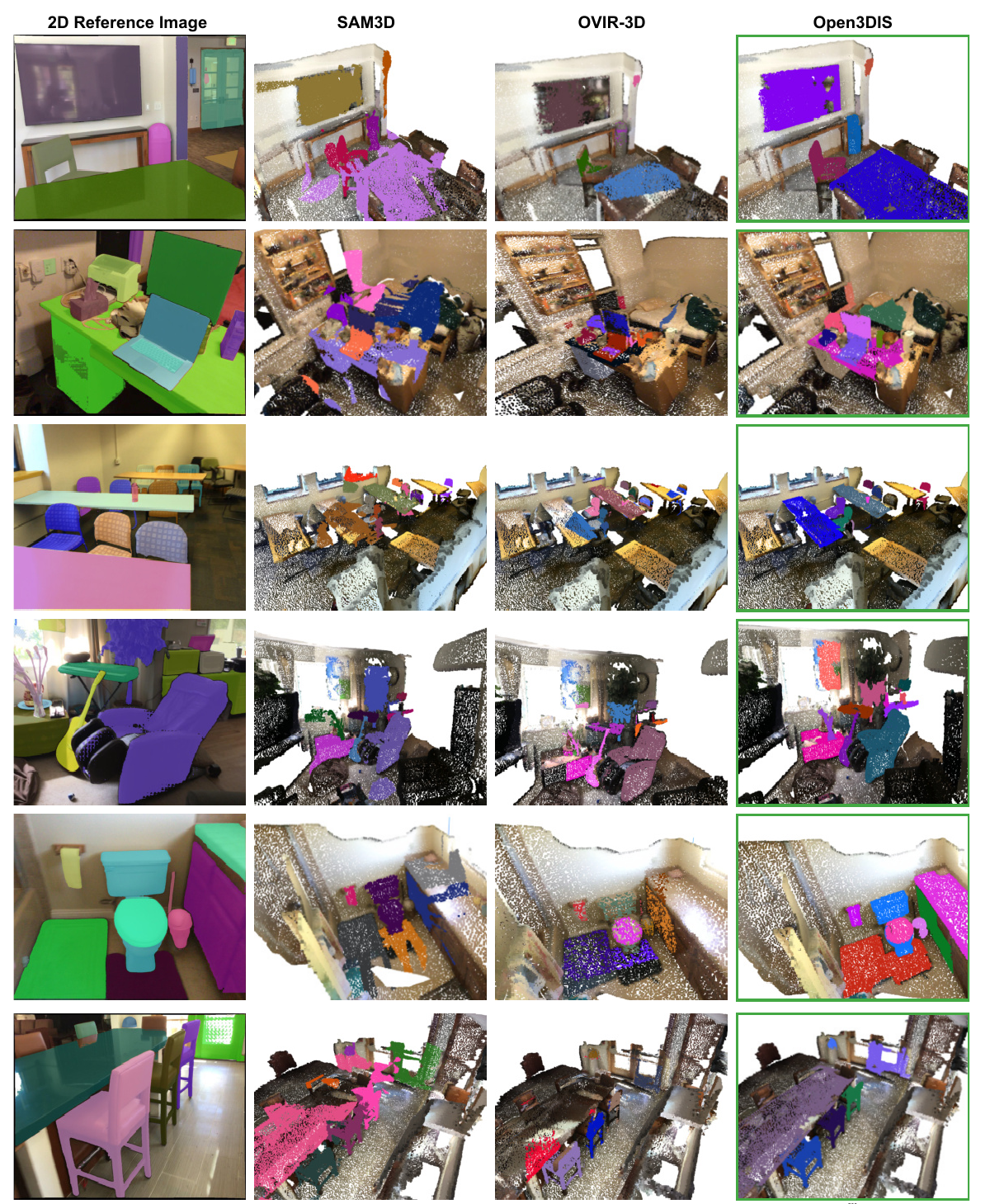}
   \caption{Qualitative results of our method compared to others in Constructing 3D proposals from 2D masks of an image. Each row shows one example, including the input 2D reference image, other 2D lifting methods, and our Open3DIS (\textbf{only 2D}) (Best viewed in color). 
   }
   \label{fig:quali_single_compare}
   \vspace{-10pt}
\end{figure*}

In order to acquire high-quality 3D augmented proposals, it is essential to guarantee the effective elevation of 2D masks from a single image to a 3D scene. The extensive overlap of 2D masks often covering multiple objects and the sensitivity of pairing points with pixels due to imperfect camera calibration are the main factors contributing to the poor performance of prior point-based approaches that rely solely on geometric Intersection over Union (IoU). In Fig. \ref{fig:quali_single_compare}, SAM3D \cite{sam3d} masks are dispersed over a wide area, while OVIR-3D \cite{ovir3d} masks are noisy and fragmented into parts. Open3DIS, however, addresses these issues by considering the superpoints 
and merging them using averaged 
3D deep features. Our method achieves consistency in 3D and 2D, yielding significantly cleaner 3D point cloud regions of corresponding masks on a single 2D image.

\subsection{Reason for Using Superpoints in 2D-G-3DIP}
\label{sec:2d-g-3d}
\begin{figure*}[t]
  \centering
\includegraphics[width=0.85\linewidth, height=0.75\paperheight]{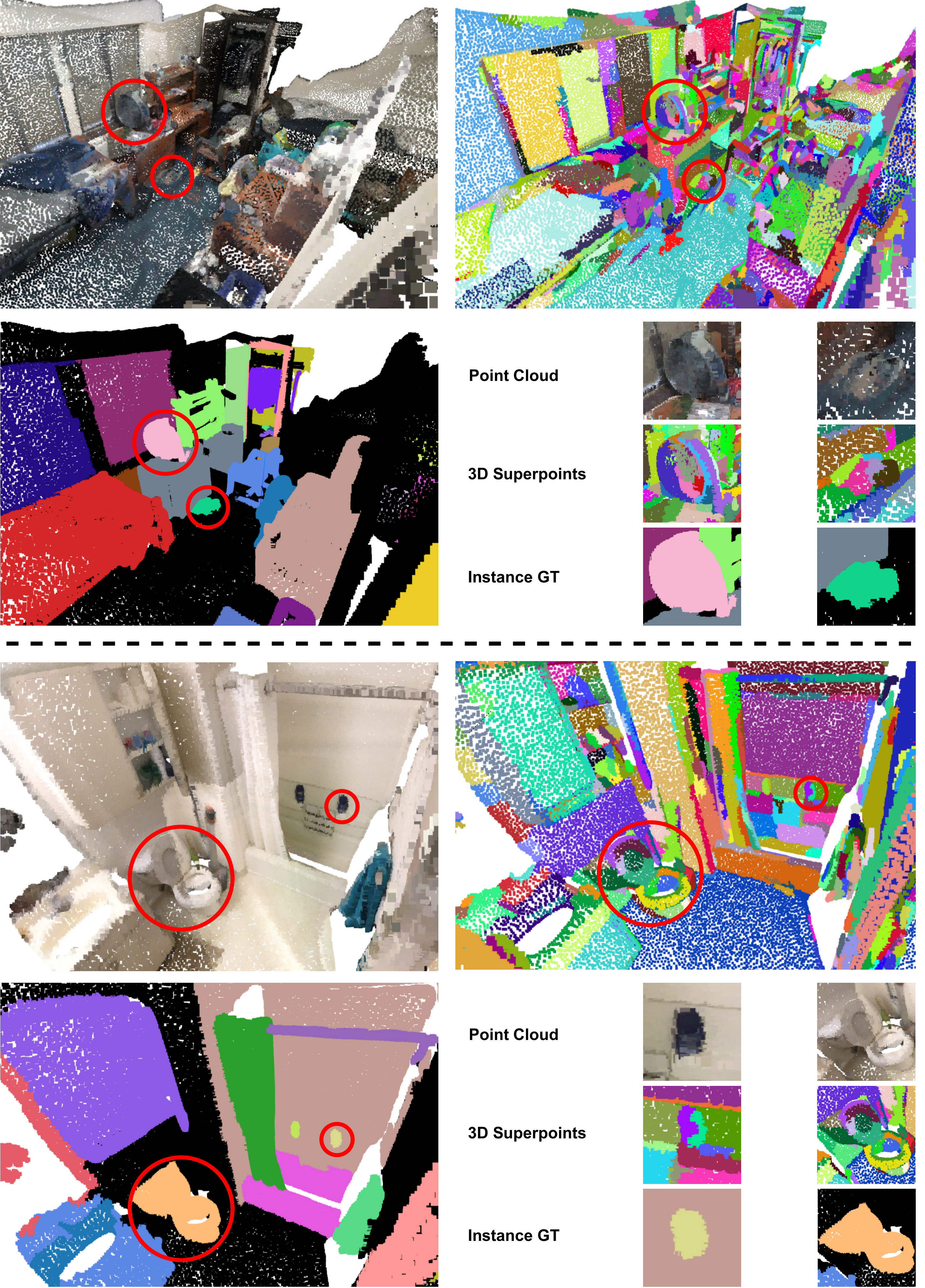}
   \caption{Two examples (separated by the dashed line) illustrating the reason for using the 2D-G-3DIP module when creating point cloud regions, with a focus on accurately covering object instances indicated by the Red circles (Best viewed in color).}
   \label{fig:sppwhy}
\end{figure*}

We have opted to utilize 3D Superpoints as the representation for our innovative 2D-G-3DIP module. The choice of 3D Superpoints is motivated by their remarkable ability to precisely encapsulate the shape and boundary of objects within a 3D scene. Essentially, when we examine an object within the 3D environment, we find that a subset of 3D Superpoints can accurately and completely cover that object's shape, as visually demonstrated in Fig. \ref{fig:sppwhy}.

Despite the potential imperfections introduced by Depth sensors, previous methods \cite{sam3d, ovir3d} have typically relied on Point Cloud - Image Projection techniques to generate \textit{Point-wise 3D instance masks}. However, this approach often yields a sparse set of 3D proposals, and some points may be obscured, resulting in incomplete masks see in Fig. \ref{fig:qualitative_compare}.

In contrast, our Open3DIS takes a distinct approach. We assign weights to groups of points, specifically 3D Superpoints, and harness the power of 3D deep features and geometric Intersection over Union (IoU) calculations. This unique combination allows us to produce \textit{Superpoint-wise 3D instance masks} that are significantly more detailed and precise than what previous methods could achieve. These masks offer a finer-grained representation of object instances in 3D scenes, even in the presence of occlusions and imperfections.


\subsection{More Qualitative Results on ScanNet200, Replica, and S3DIS}
\begin{figure*}[t]
  \centering
\includegraphics[width=1\linewidth]{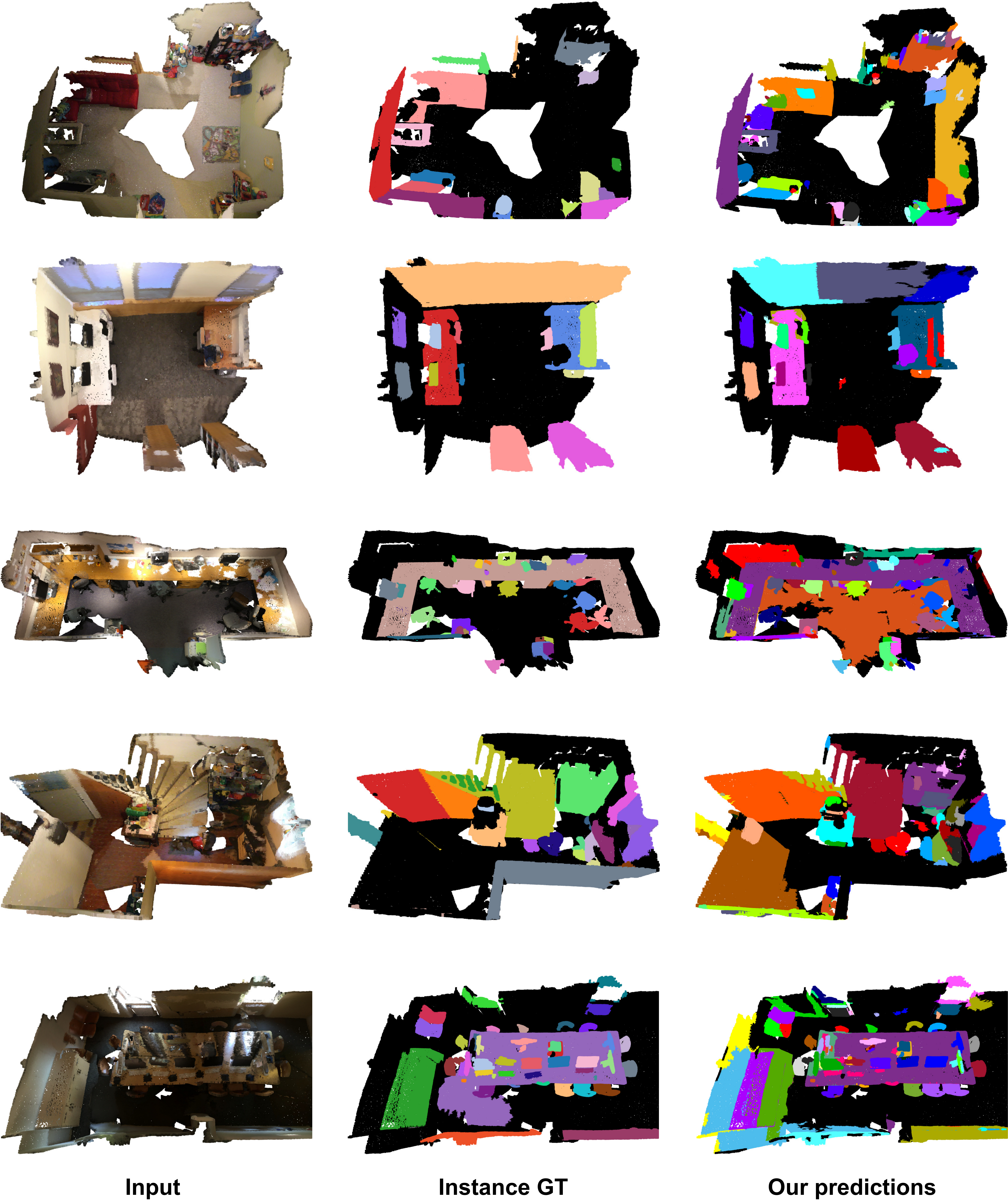}
   \caption{Qualitative results of our method on the ScanNet200 dataset. Each row shows one example, including the input RGB point cloud, instance ground truth, and our predictions (Best viewed in color).}
\label{fig:qualitative_scannet200}
   \vspace{-10pt}
\end{figure*}

\myheading{ScanNet200.}
We present visualizations of Open3DIS applied to the extensive Scannet200 dataset. In Fig.~\ref{fig:qualitative_scannet200}, we display scenes that have been processed by Open3DIS alongside their corresponding Instance Ground Truth (Instance GT). Despite the considerable size of the Scannet200 dataset, it is important to note that the ground truth annotations may overlook certain relatively small objects within the scenes. These omitted objects are represented by black points, indicating instances that have not been labeled.
Open3DIS utilizes both 2D and 3D segmenters to generate comprehensive 3D instance masks, ensuring that even significantly small objects are covered. Although we continue to use the Scannet200 dataset for evaluation purposes, primarily due to its inclusion of a wide range of object classes, we anticipate that Open3DIS will demonstrate notably superior performance when applied to finer-grained 3D instance segmentation datasets.

\begin{figure*}[t]
  \centering
\includegraphics[width=1\linewidth]{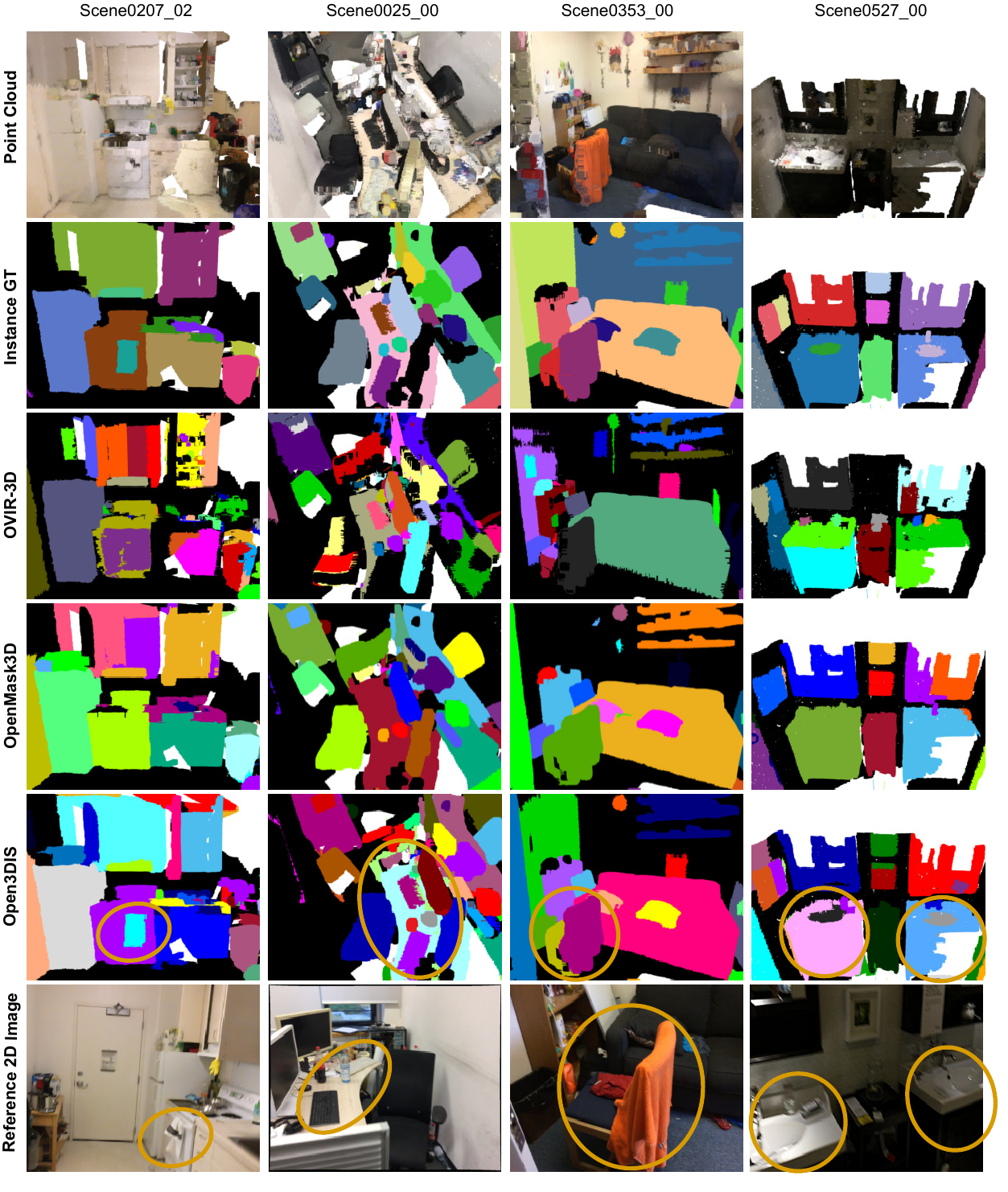}
   \caption{Qualitative results of our method compared to others on ScanNet200 dataset. Each column shows one example in Orange ellipses demonstrating that \Approach~performs better than others (Best viewed in color).}
   \label{fig:qualitative_compare}
   \vspace{-10pt}
\end{figure*}

In comparison to other methods, as depicted in Fig.~\ref{fig:qualitative_compare} with a closer look, Open3DIS excels in producing finer 3D masks that effectively cover objects with complex and ambiguous geometric structures. On the other hand, OVIR-3D relies on 2D segmenters and directly extends 2D masks to 3D scenes through point-based Intersection over Union (IoU) matching. This approach results in suboptimal mask quality, despite its capability to discover rare object classes.
In contrast, OpenMask3D employs a 3D instance segmenter and evaluates each 3D instance using the CLIP model. While this approach may offer benefits in certain scenarios, it compromises the generality of Open-Vocabulary 3D Instance Segmentation (Open-Vocabulary 3DIS). Particularly, OpenMask3D may struggle to identify rare object classes when expanding the number of classes during training.

Tab. 3 in the main paper provides an illustration of these differences. OpenMask3D, when trained on Scannet20, achieves an Average Precision (AP) score of 12.6, whereas Open3DIS surpasses the state-of-the-art method with an impressive AP score of 19.0. This substantial performance gap underscores Open3DIS's superiority in handling diverse and challenging 3D instance segmentation tasks.

\begin{figure*}[h]
    \centering
    \begin{subfigure}[b]{.9\textwidth}
        \centering
        \includegraphics[width=1\linewidth]{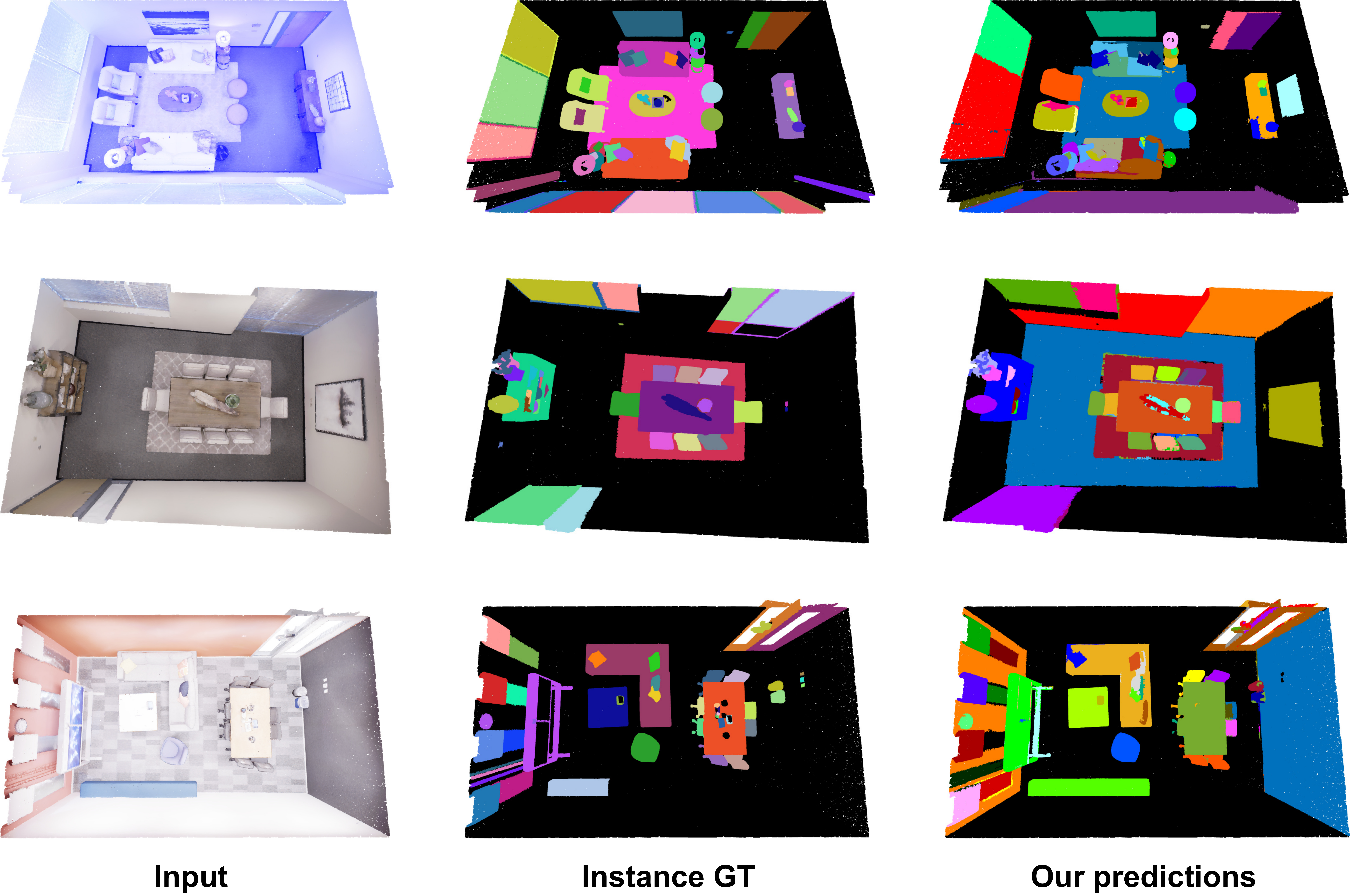}
        \caption{Qualitative results on Replica}
        \label{fig:qualitative_replica}
    \end{subfigure}
    
    \begin{subfigure}[b]{.9\textwidth}
        \centering
        \includegraphics[width=1\linewidth]{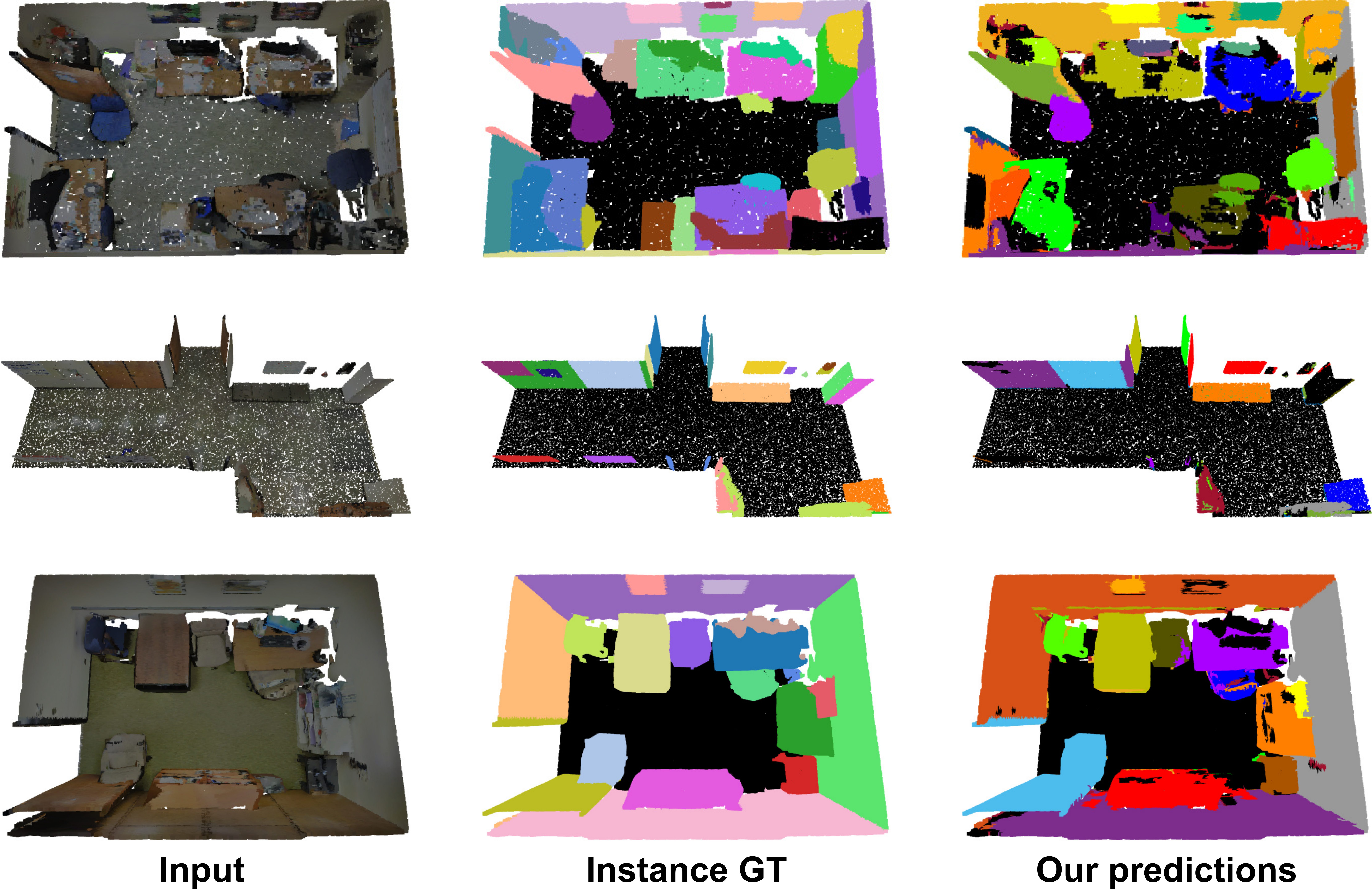}
        \caption{Qualitative results on S3DIS}
        \label{fig:qualitative_s3dis}
    \end{subfigure}
    \caption{Qualitative results of our method on the Replica (Top) and S3DIS (Bottom) datasets. Each row shows one example, including the input RGB point cloud, instance ground truth, and our predictions (Best viewed in color).}
    \label{fig:both_images}
\end{figure*}

\myheading{Replica.}
The qualitative results of our approach on the Replica dataset are visualized in Fig.~\ref{fig:qualitative_replica}.

\myheading{S3DIS.}
The qualitative results of our approach on the S3DIS dataset are visualized in Fig.~\ref{fig:qualitative_s3dis}.

\subsection{Open-Vocabulary Scene Exploration}
\begin{figure*}[h]
    \centering
    \begin{subfigure}[b]{0.49\textwidth}
        \centering
        \includegraphics[width=1\textwidth]{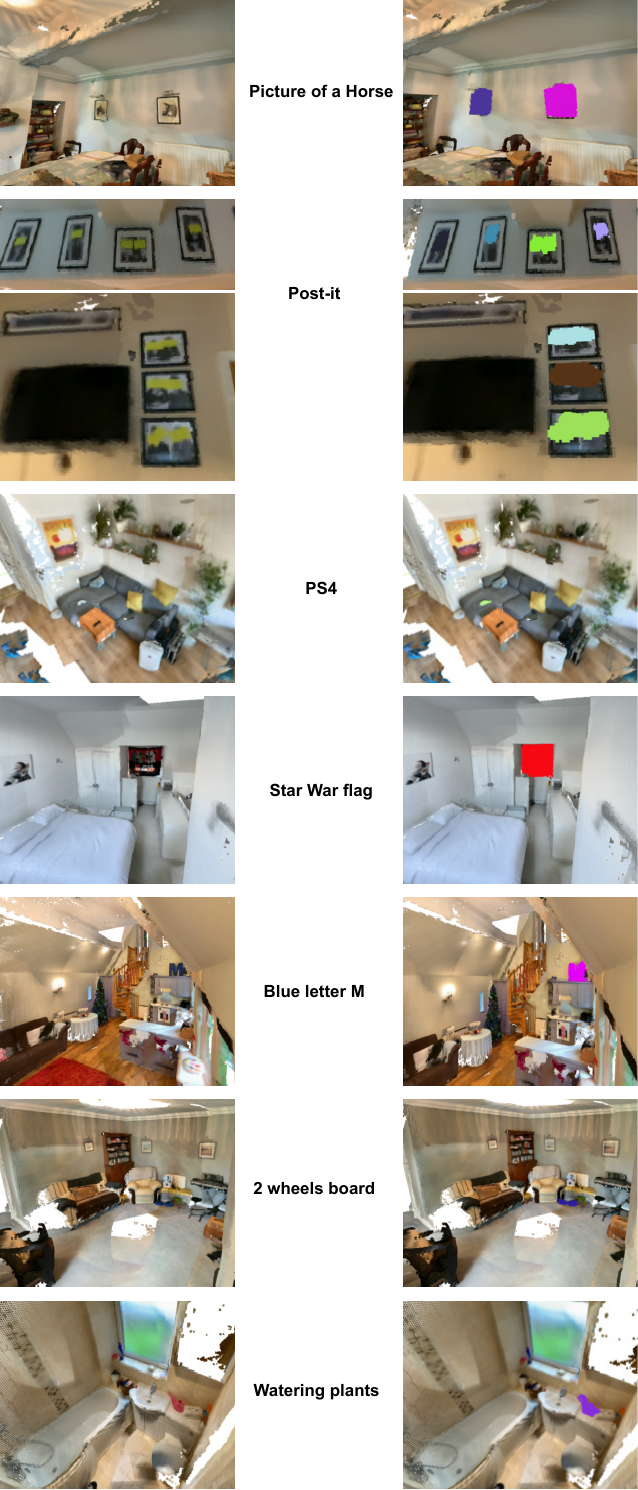}
        \vspace{1pt}
        \caption{ARKitScenes}
        \label{fig:ov-arkit}
    \end{subfigure}
    \hfill
    \begin{subfigure}[b]{0.49\textwidth}
        \centering
        \includegraphics[width=\textwidth]{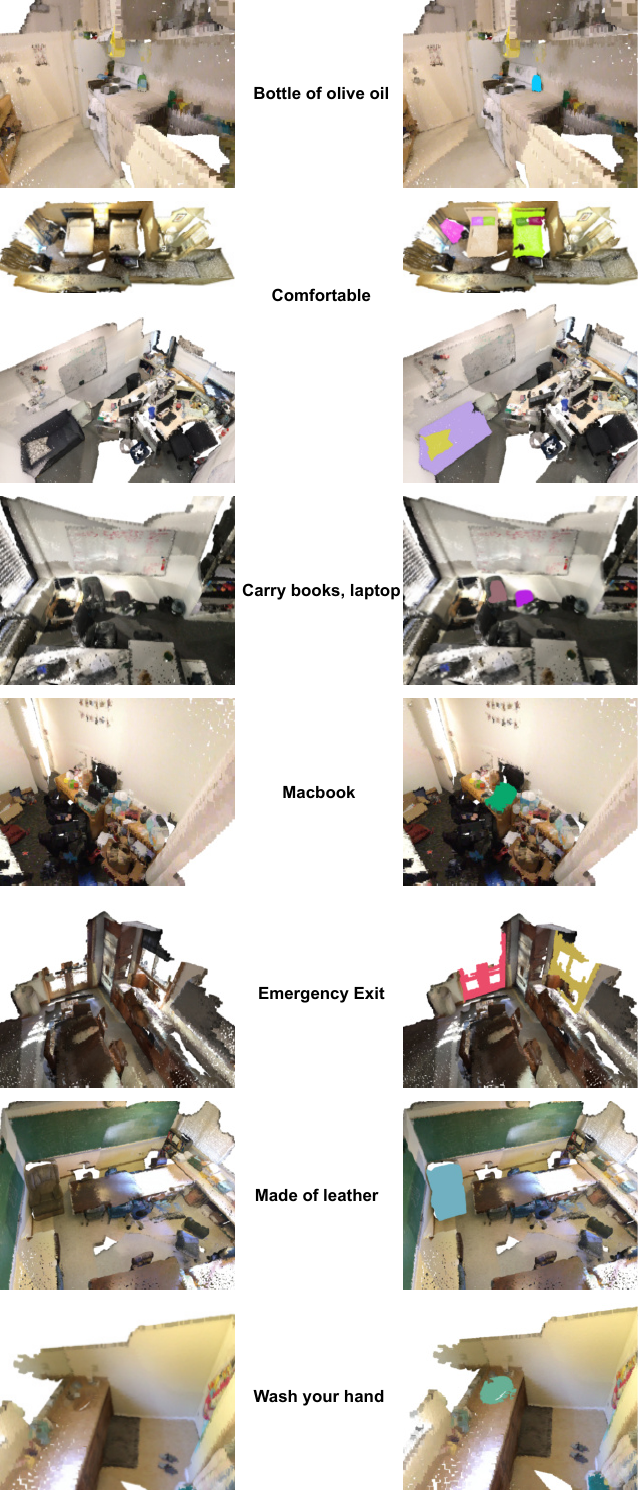}
        \vspace{1pt}
        \caption{Scannet200}
        \label{fig:ov-scannet}
    \end{subfigure}
    \caption{Open-Vocabulary exploration on \textbf{ARKitScenes} \cite{dehghan2021arkitscenes} (Left) and \textbf{Scannet200} \cite{scannet200} (Right) with Open3DIS (2D only). The middle column presents the text queries,  the original point cloud is displayed on the left column, and colored regions represent 3D instance proposals on the right column. (Best viewed in color, zoom-in is advised).}
    \label{fig:both_images}
\end{figure*}

We showcase the remarkable Open-Vocabulary scene exploration capabilities of Open3DIS on the ARKitScenes \cite{dehghan2021arkitscenes} (Fig.~\ref{fig:ov-arkit}) and ScanNet200 \cite{scannet200} (Fig.~\ref{fig:ov-scannet}) datasets, which are notable for containing a vast array of scenes featuring diverse and rare objects. Specifically, we demonstrate the system's ability to query instance objects based on various attributes such as material, color, affordances, and usage. We intentionally exclude the Class-agnostic 3D Segmenter component, thereby pushing our method toward a near Zero-Shot Instance Segmentation approach.
Remarkably, in challenging scenarios, such as identifying objects like a Post-it note, a picture of a horse, or a bottle of olive oil, Open3DIS outperforms other methods \cite{openmask3d, ovir3d, sam3d, openscene} significantly. Some of these methods struggle to detect these objects, let alone locate them accurately.
%
%
%
\textit{Please see the supplementary video for a live demo.}



%
%






\clearpage

\end{document}